\pdfoutput=1

\documentclass[11pt]{article}

\usepackage[]{acl}

\usepackage{times}
\usepackage{latexsym}
\usepackage{microtype}
\usepackage{amssymb}
\usepackage{amsmath}
\usepackage{bbm}
\usepackage{inconsolata}
\usepackage{float}
\usepackage{multirow}
\usepackage{graphicx}
\usepackage{booktabs}
\usepackage{caption}
\usepackage{subcaption}
\usepackage{bbold}
\usepackage{xspace}
\usepackage{makecell}
\usepackage[capitalize,noabbrev]{cleveref}

\usepackage[T1]{fontenc}

\usepackage[utf8]{inputenc}

\newcommand{\selfinstruct}{\textsc{Self-Instruct}\xspace}
\newcommand{\abspyramid}{\textsc{AbsPyramid}\xspace}
\newcommand{\papertitle}{\textsc{AbsInstruct}\xspace}
\newcommand{\textitbf}[1]{\textbf{\textit{#1}}}
\newcommand{\nounrelation}{\textit{Noun-Entail}\xspace}
\newcommand{\verbrelation}{\textit{Verb-Entail}\xspace}
\newcommand{\eventrelation}{\textit{Event-Entail}\xspace}
\definecolor{stepcolor}{HTML}{dd8452}
\definecolor{contentcolor}{HTML}{55a868}
\newcommand{\textttbf}[1]{\texttt{\textbf{#1}}\xspace}
\title{\papertitle: Eliciting Abstraction Ability from LLMs through\\ Explanation Tuning with Plausibility Estimation}

\author{
  Zhaowei Wang$^1$, Wei Fan$^1$, Qing Zong$^1$, Hongming Zhang$^2$, Sehyun Choi$^1$, \\ \textbf{Tianqing Fang$^1$, Xin Liu$^3$, Yangqiu Song$^1$, Ginny Y. Wong$^4$, \& Simon See$^4$} \\
  $^1$Department of Computer Science and Engineering, HKUST\\ $^2$Tencent AI Lab, Bellevue, USA, $^3$Amazon.com Inc, Palo Alto, USA\\
  $^4$NVIDIA AI Technology Center (NVAITC), NVIDIA, Santa Clara, USA\\
  \texttt{\{zwanggy, yqsong\}@cse.ust.hk,} \texttt{\{gwong, ssee\}@nvidia.com}
  }

\begin{document}
\maketitle
\begin{abstract}
Abstraction ability is crucial in human intelligence, which can also benefit various tasks in NLP study. Existing work shows that LLMs are deficient in abstract ability, and how to improve it remains unexplored. In this work, we design the framework \papertitle to enhance LLMs' abstraction ability through instruction tuning. The framework builds instructions with in-depth explanations to assist LLMs in capturing the underlying rationale of abstraction. Meanwhile, we introduce a plausibility estimator to select instructions that are more consistent with the abstraction knowledge of LLMs to be aligned. Then, our framework combines abstraction instructions with general-purpose ones to build a hybrid dataset. Extensive experiments and analyses\footnote{The code and data are available at \url{https://github.com/HKUST-KnowComp/AbsInstruct}} demonstrate that our framework can considerably enhance LLMs' abstraction ability with strong generalization performance while maintaining their general instruction-following abilities.

\end{abstract}

\section{Introduction}
Abstraction ability is central to human
cognition~\cite{minsky1980k}, which is identifying shared traits among items to build a broader concept, like deriving the concept of ``beverage'' from ``coffee'' and ``tea.'' With this ability, we can derive general rules and principles from past experiences, which enables us to adeptly navigate new situations in our daily life~\cite{russell2010artificial,abstraction2013saitta}. In NLP, building abstraction resources has long been a vital challenge to which the community has devoted many efforts~\cite{ hosseini2018learning,he2022acquiring}.

\begin{figure}[t]
    \centering
    \includegraphics[width=\columnwidth]{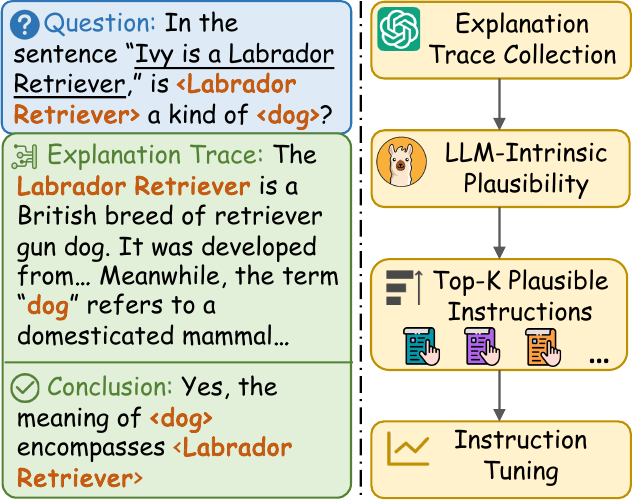}
    \caption{An illustration of our \papertitle framework. We collect explanation traces for each example and design a plausibility estimator to select data that match the knowledge of an LLM to be aligned.}
    \label{fig:intro_illustration}
\end{figure}

Among them, \citet{wang2023abspyramid} built the first comprehensive benchmark, \abspyramid, of abstract concepts for nouns, verbs, and events. In this benchmark, models are asked to detect the validity of an abstract concept, as shown in \cref{fig:intro_illustration}. Their evaluations on the benchmark reveal that abstraction remains challenging even for state-of-the-art LLMs. For example, ChatGPT~\cite{openai2023chatgpt} only modestly exceeds majority voting and substantially trails behind fine-tuned smaller models. While prior works have explored ways for general-domain LLM alignment~\cite{sanh2021multitask, ouyang2022training}, how to elicit the abstraction knowledge of LLMs remains unexplored.

Nonetheless, enhancing LLMs' abstraction ability is a non-trivial task. We only observe slight improvements when gathering vanilla instructions from randomly sampled data for detecting abstract concepts. First, the responses of vanilla instructions only express the validity of abstract concepts as ``Yes/No.'' As a result, LLMs might only grasp the surface-level styles but miss underlying rationales in deciding the validity of abstract concepts~\cite{kung2023models}. Moreover, existing studies show that LLMs acquire most of the knowledge and abilities during pre-training~\cite{zhou2023lima,jha2023limit}. Thus, instructions from randomly sampled data might not be consistent with the abstraction knowledge of pre-trained models for better elicitation.

To tackle those issues, we propose the framework \papertitle to build instructions with detailed explanation traces and well-crafted data selection, as shown in \cref{fig:intro_illustration}. The framework forms explanation traces by collecting meanings of each given instance and abstract concept. These traces can help LLMs better comprehend the underlying reasoning process of detecting abstract concepts. Moreover, we introduce a plausibility estimator to select instruction data consistent with the abstraction knowledge of a pre-trained model to be aligned. The estimator assesses the plausibility score of each example based on the probability computed by the pre-trained model. Then, we only retain examples with higher plausibility scores, which align better with the model's knowledge. We also introduce a collection of filters based on lexical overlap, keywords, and predicted labels to ensure diversity and quality further. Ultimately, a hybrid dataset is constructed by combining instructions for abstraction detection with those in the general domain.

For evaluation, the framework first builds instructions for abstraction detection based on \abspyramid~\cite{wang2023abspyramid} and combines them with instructions from Alpaca~\cite{taori2023alpaca}. Next, we conduct extensive experiments and analyses of several popular LLMs instruction-tuned with our framework. The evaluation results show that applying \papertitle can effectively unlock LLMs' abstraction ability, with the performance surpassing existing alignment methods by a large margin of 6-10\%. Also, thorough ablation studies corroborate the efficacy of explanation traces, the plausibility estimator, and various filters. Meanwhile, we conduct detailed analyses to show the robustness of our framework and the generalization ability of LLMs trained with our framework. Last but not least, the automatic and human evaluations on two general-domain instruction datasets, SuperNI~\cite{wang2022super} and \selfinstruct~\cite{wang2023self}, manifest that our framework can enhance abstraction ability without compromising LLMs' performance of following general instructions.

\section{Related Work}
Abstraction has long been widely applied across various tasks, including question answering~\cite{zheng2023take}, machine translation~\cite{pado2009measuring}, and many others~\cite{yoshikawa2019combining,khot2018scitail,mckenna2021multivalent}. While some works have studied entity abstraction~\cite{clark2000exploiting,wu2012probase,song2015open} without considering contexts, our work explores event abstraction with a few relevant fields:

\paragraph{Event Abstraction:} 
\label{sec:related_work_abs}
This field focuses on studying abstraction within an event as context. One line of works studied extracting entailment graphs for verbs with two arguments from large corpora~\cite{berant2011global,hosseini2018learning, hosseini2019duality, hosseini2021open,guillou2020incorporating,chen2022entailment,mckenna2021multivalent,mckenna2023smoothing}. Meanwhile, \citet{he2022acquiring} curated abstract concepts for nouns and events based on ATOMIC~\cite{sap2019atomic}. Recently, \citet{wang2023abspyramid} compiled a large benchmark that unifies the scopes of the abovementioned works. They collected abstraction descriptions of events and hypernyms of nouns and verbs using ChatGPT~\cite{openai2023chatgpt} and WordNet~\cite{miller1995wordnet}, which are then manually annotated. Their studies suggest that LLMs still struggle with abstraction knowledge with various mistakes. Thus, we present the first attempt to unlock the stronger abstraction abilities of LLMs.

\begin{figure*}[t]
    \centering
    \includegraphics[width=2\columnwidth]{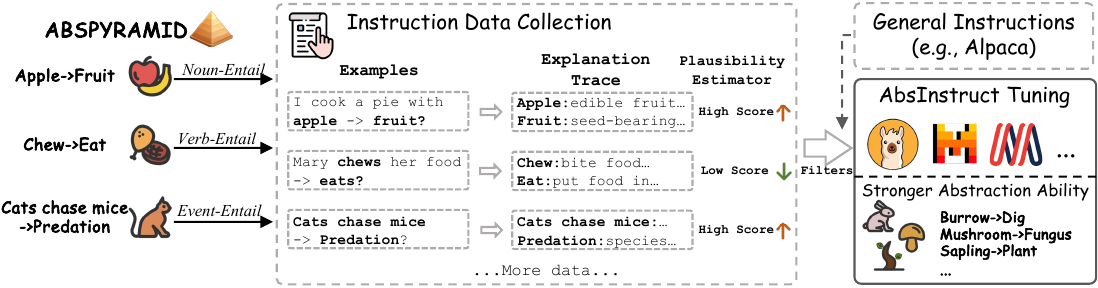}
    \caption{The overview of our \papertitle framework. We sample examples from \abspyramid and collect explanation traces by prompting an LLM. Then, we design a plausibility estimator to choose examples that are more consistent with the knowledge of a model to be aligned. The framework combines abstraction instructions with general-domain ones (e.g., Alpaca) and instruction-tunes the model.}
    \label{fig:method_figure}
\end{figure*}

\paragraph{Linguistic Entailment:} In linguistics, the studies of event abstraction are guided by the concept of linguistic entailment~\cite{beth1955semantic,murphy2010lexical,indarti2015distinguishing}, which is enforced by lexical semantics combined with the laws of logic. For example, \textit{Bella is a friendly kitten} entails \textit{Bella is a cat}, as one cannot be a friendly kitten without being a cat. Importantly, linguistic entailment contrasts with textual entailment~\cite{dagan2005pascal}, also called NLI~\cite{bowman2015large,conneau2018xnli}, which emphasizes what \textitbf{typicially} can be inferred from a premise and can be fallible.

\paragraph{Instruction Tuning:} Aligned LLMs are strongly preferred by humans over original ones~\cite{zheng2023judging,chiang2023vicuna}, and diverse methods are studied to curate instructions, such as NLP tasks~\cite{mishra2022cross, chung2022scaling}, real user requests~\cite{conover2023free}, and synthetic instructions~\cite{wang2023self}. Recent studies~\cite{mukherjee2023orca,mitra2023orca} suggest that instructions with detailed responses can provide underlying rationales and enhance alignment efficacy. Meanwhile, several works~\cite{zhou2023lima, jha2023limit, song2023dynamics} demonstrate that an LLM captures almost all the knowledge during pre-training, which can be unlocked even with a small number of instructions during alignment. Motivated by these discoveries, we collect abstraction instructions with explanation traces and try to select instructions more consistent with LLMs' knowledge for better elicitation.

\section{Method}
Eliciting abstraction knowledge from pre-trained LLMs can be challenging since it requires (1) underlying rationales in determining the validity of abstract concepts and (2) carefully curated instructions to better elicit the knowledge. Here, we describe the process of \papertitle, which builds instructions with explanation traces and employs a plausibility estimator and several filters for data selection. This pipeline is depicted in Figure~\ref{fig:method_figure}.

\subsection{Data Format Definition}
Our work concentrates on detecting valid abstract concepts~\cite{wang2023abspyramid}, defined as a binary classification task. The task input is a five-element tuple in the format of (\textitbf{head event}, \textitbf{entailment relation}, \textitbf{tail event}, \textitbf{instance}, \textitbf{concept}). In detail, the \textitbf{instance} is a component of the \textitbf{head event}, which can be a noun, verb, or entire event. Then, we replace the instance with its \textitbf{concept} to build the \textitbf{tail event}. Models are asked to decide whether the concept is a valid abstraction of the instance, where the head event linguistically entails the tail event. Here, we study three \textitbf{entailment relations} defined on instance types: \nounrelation, \verbrelation, and \eventrelation. We provide concrete examples in \cref{app:filtered_example,app:case_study}.

The format of instruction data consists of
three elements: \textitbf{instruction}, \textitbf{input}, and \textitbf{response}. An \textitbf{instruction} outlines the task using natural language while an \textitbf{input} and \textitbf{response} serve as a task example. 
Note that the input is optional because of the blurred boundary between it and the instruction. 
For example, while ``Give me a report about the following topic'' and ``global economics'' can serve as separate instruction and input, we also can combine them as a sole instruction: ``Give me a report about global economics.''

\subsection{Instruction and Input Compilation}
We manually build instructions for all entailment relations: \nounrelation, \verbrelation, and \eventrelation. Since our framework introduces detailed responses with explanation traces, the instructions for each relation comprise two steps, asking LLMs to (1) consider the meanings of the given instances and concepts and (2) predict the label based on the explanation in the first step. 

Next, we collect the input of abstraction detection for each relation. Our framework samples five-element tuples with balanced labels from the training set of \abspyramid~\cite{wang2023abspyramid}. To build the input, we verbalize each tuple using prompts that ask whether the concept is a valid abstraction of the instance, given the head event as context.
We provide concrete prompts for building the instructions and input in \cref{app:abs_instruct_prompt}.

\subsection{Response Collection with Explanation}
In conformity with instructions, our framework collects responses consisting of two steps: (1) the \textitbf{explanation step}, which contains the meanings of given words, and (2) the \textitbf{conclusion step}, which confirms the concept validity by comparing word meanings. The easy-to-build component is the \textitbf{conclusion step}. For each example, we verbalize the binary label as ``Yes'' or ``No'' and append a short comparison, such as ``Yes, the meaning of \textbf{[cpt]} encompasses \textbf{[ins]},'' where \textbf{[ins]} and \textbf{[cpt]} are two placeholders for the given instance and concept.

For the \textitbf{rationale step}, we first conduct a pilot study about using taxonomies to build explanation traces, such as WordNet~\cite{miller1995wordnet}, which can provide meanings of nouns and verbs. Our findings disclose two problems with using a taxonomy. First, the coverage of nouns in WordNet is inadequate. Only 6.32\% of nominal phrases can be found in WordNet. For example, while ``cat'' is incorporated in WordNet, many specific types are absent, such as fluffy cat and ginger cat. Moreover, we need word sense disambiguation~\cite{pradhan2007semeval} to choose correct word meanings, which may accumulate errors in our framework. For example, the expert annotation shows that only 61.0\% of WSD results from GlossBERT~\cite{huang2019glossbert} are correct (Details in \cref{app:wsd_annotation}).

To overcome those challenges, we build explanation traces with the help of an LLM. In detail, we prompt GPT4 under the zero-shot setting with the instruction asking the meaning of a given word. We collect the meanings of the instance and concept separately and then concatenate them to build the whole explanation trace. After collecting both steps, the framework constructs the whole response with the format:
\begin{center}
\begin{tabular}{l}
\noindent\textttbf{\textcolor{stepcolor}{Step1: }\textcolor{contentcolor}{<ins mean>} Meanwhile, \textcolor{contentcolor}{<cpt mean>}}\\
\noindent\textttbf{\textcolor{stepcolor}{Step2: }\textcolor{contentcolor}{Yes/No, the meaning of ...}}
\end{tabular}
\end{center}
where \textttbf{<ins mean>} and \textttbf{<cpt mean>} stand for the meanings of the instance and concept. The whole response interprets and compares the given instance and concept, assisting LLMs in seizing the underlying reasoning processes. We provide concrete prompts for using GPT4 in \cref{app:word_meaning_prompt}.

\subsection{Example Postprocessing} 
After gathering many examples, we employ several filters and a plausibility estimator to select instructions. First, two quality filters are introduced to remove basic errors: the prediction filter and the keyword filter. Then, we introduce a diversity filter based on ROUGE-L~\cite{lin2004rouge} to remove similar examples. Lastly, we design a plausibility estimator to select abstraction examples consistent with pre-trained LLMs' knowledge.

\begin{figure}[t]
    \centering
    \vspace{-0.2in}
    \includegraphics[width=0.8\linewidth]
    {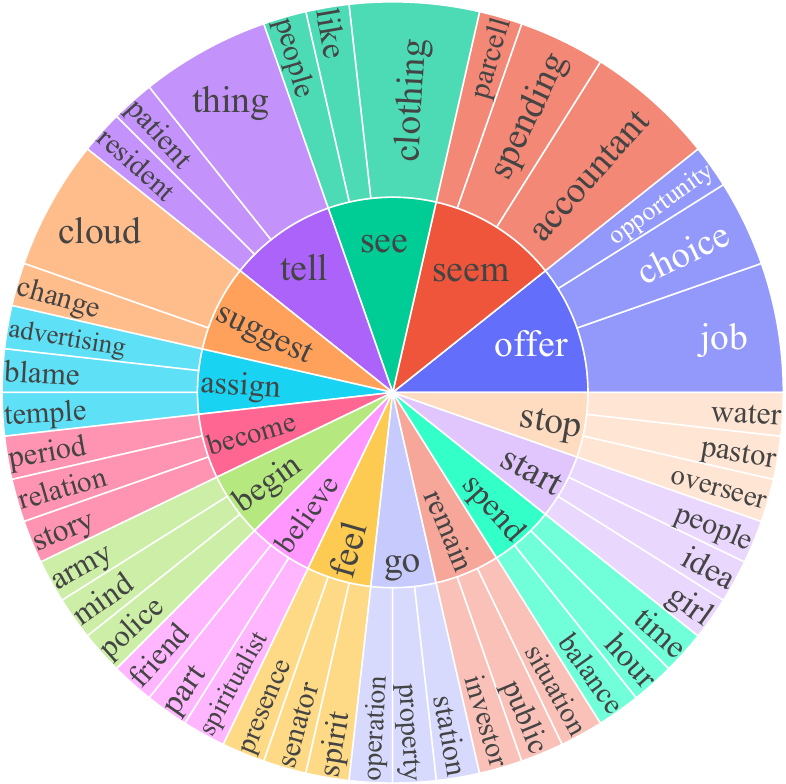}
    \caption{The 15 most common verbs (inner circle) and their top 3 direct nominal objects (outer circle) in head events of collected examples.}
    \label{fig:verb_noun_pie}
\end{figure}

\paragraph{Prediction Filter:} A faithful explanation trace should assist LLMs in reaching the correct prediction. Therefore, given the explanation trace we built, we prompt GPT4 to predict a label for each example. Then, we discard all examples that GPT4 cannot give the right answer:

\begin{equation}\begin{aligned}
\begin{cases}
            \hat{y} = \theta_{LLM}(i, x, e) \\
            f_{\textit{pred}}(\hat{y}, y) = \mathbbm{1} \left\{ \hat{y} = y \right\},
\end{cases}\end{aligned}\end{equation}
where $\theta_{LLM}$ signifies the parameters of GPT4 that outputs a predicted label $\hat{y}$ given the instruction $i$, input $x$, and explanation trace $e$. Then, the filter $f_{\textit{pred}}$ compares $\hat{y}$ with ground truth $y$. 

\paragraph{Keyword Filter:} We observe that GPT4 may explain the meaning of another word in the head event rather than the given one due to hallucination (See cases in \cref{app:filtered_example}). Thus, we design the keyword filter to discard examples whose explanation trace omits its instance or concept. Take \cref{fig:intro_illustration} as an example. The explanation must contain both the keywords ``Labrador Retriever'' and ``dog.''

\paragraph{Diversity Filter}
Our framework collects a large pool of examples from \abspyramid, which could result in multiple examples with similar instances or concepts. To promote diversity, a new example is added only if its ROUGE-L similarity with any existing example is below 0.7, following prior works~\cite{wang2023self, taori2023alpaca}.

\paragraph{Plausibility Estimator} Existing studies~\cite{zhou2023lima, jha2023limit} show that a model obtains its knowledge almost entirely during pre-training, which can be elicited with a modest set of examples during alignment. For better elicitation, we select examples that are more consistent with the knowledge of the pre-trained LLM to be aligned. Here, we measure the LLM-intrinsic plausibilities of each example, which is determined by the model's knowledge. Concretely, the plausibility is computed as the normalized conditional probability of the response $r$ given the instruction $i$ and input $x$: 
\begin{equation}\begin{aligned}
            Plausibility(i, x, r) = P_{\theta}(r | i, x)^{\frac{1}{N}},
\end{aligned}\end{equation}
where $\theta$ are the parameters of the pre-trained model, and $N$ is the number of tokens in $r$. The above equation is equivalent to the reciprocal of the perplexity of $r$ conditional on $i$ and $x$. Then, the framework only retains examples with top-$K$ plausibilities. Note that we compute plausibilities based on a model's intrinsic knowledge, in contrast to those definitions on real-world knowledge~\cite{wu2012probase,chalier2020joint}. In practice, we take the logarithm of the above equation to ensure numerical stability.

\subsection{Mixed Alignment Data}
Our framework combines the abstraction instructions we collected and general-domain instructions to build the final dataset. The dataset is then used to finetune the same model that we use to compute plausibility scores. We concatenate an instruction and the input as a prompt (See details in \cref{app:prompt_for_concatenation}) and train the model to generate the response in a standard supervised way.

\section{Abstraction Instruction Overview}
In this section, we apply \papertitle for inducing instruction data as a case study, with Llama2 (7B)~\cite{touvron2023llama} used to estimate plausibilities. Our framework constructs 200 examples for each relation, derived from \abspyramid.

\begin{figure}[t]
    \centering
    \includegraphics[width=\columnwidth]{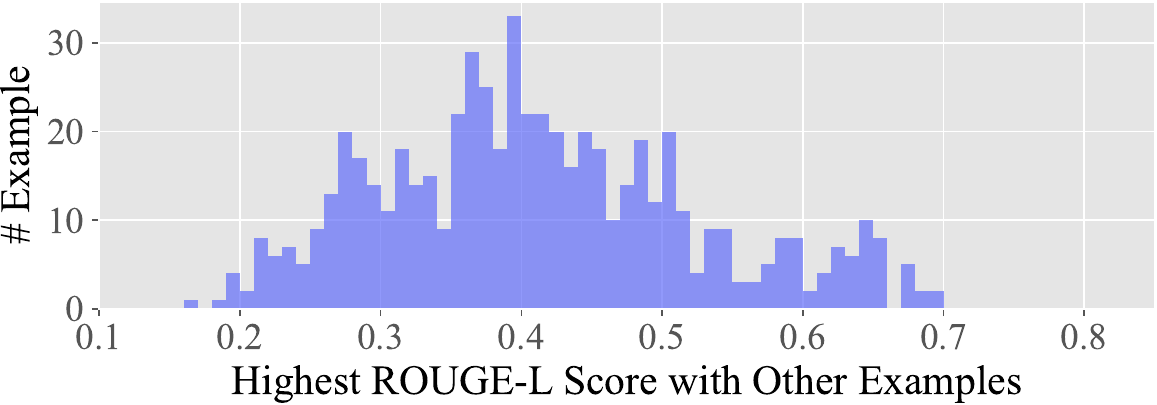}
    \caption{Distribution of the ROUGE-L scores
between collected examples. For each example, we compute the highest similarity with other examples we gathered.}
    \label{fig:diversity_llama2_7b}
\end{figure}

\subsection{Diversity}
We identify the verb-noun structure in the head events of examples to examine the diversity of collected examples. We use the Berkeley Neural Parser~\cite{kitaev2018constituency, kitaev2019multilingual} to parse each event and then extract the verb that is closest to the root as well as its first nominal object. 391 out of 600 head events contain such structure as other events usually are more complex, such as ``\texttt{PersonX} began renting the space to businesses.'' We plot the 15 most common verbs and their top 3 direct nominal objects in \cref{fig:verb_noun_pie}, which makes up 9.67\% of the entire set. Overall, we see diverse topics and textual formats in these examples. 

We further study the diversity of collected examples. For each example we collect, we compute its highest ROUGE-L similarity with other ones. We plot the distribution of these ROUGE-L scores in \cref{fig:diversity_llama2_7b}. The results indicate a decent number of unique examples, which do not overlap much with the remaining.


\begin{table}[t]
\small
\setlength\doublerulesep{\arrayrulewidth}
 \renewcommand\cellset{\renewcommand\arraystretch{0.5}}
\centering
\begin{tabular}{lc}
\toprule
Quality Review Question & Yes \%\\ 
\midrule
\makecell[l]{Is the explanation of the instance correct?}&94.7\%  \\ \midrule
\makecell[l]{Is the explanation of the concept correct?}&96.0\%  \\ \midrule
\midrule
All fields are valid&  92.0\%\\ 
\bottomrule
\end{tabular}
\caption{Data quality annotation for explanation traces generated by GPT4.}
\label{tab:data_quality_eval}
\end{table}

\subsection{Quality}
To investigate the quality, we sample 150 examples and ask three experts to label the correctness of the meanings of instances and concepts (See details in \cref{app:quality_collected_example}). Results in \cref{tab:data_quality_eval} demonstrate that most of the collected explanation traces are meaningful. While some traces may contain noise, we found that explanation traces can provide useful guidance for tuning LLMs for abstraction ability.


\begin{table*}[t]
    \small
	\centering
	\begin{tabular}{l|l||cc|cc|cc|cc}
	\toprule
        \multirow{2}{*}{\textbf{Methods}}&\multirow{2}{*}{\textbf{Backbone}}&\multicolumn{2}{c|}{\textbf{Noun}} &\multicolumn{2}{c|}{\textbf{Verb}}&\multicolumn{2}{c|}{\textbf{Event}}&\multicolumn{2}{c}{\textbf{All}}\\ 
	&&\textbf{Acc} &\textbf{Ma-F1} &\textbf{Acc}&\textbf{Ma-F1} & \textbf{Acc}&\textbf{Ma-F1}&\textbf{Acc}&\textbf{Ma-F1} \\
            \midrule
            \textbf{Random} & \multicolumn{1}{|c||}{-} &50.00 & 49.56 & 50.00 & 49.95 & 50.00 & 48.98 & 50.00 & 49.50 \\
            \midrule
            \multirow{4}{*}{\textbf{LLM API (Zero)}}
            &GPT 4 &79.70&77.34&57.50&54.24&69.70&63.32&68.97&64.97\\
            &GPT 3.5&67.00&62.45&56.30&55.90&65.60&58.23&62.97&58.86\\
            &ChatGPT&74.00&72.27&56.30&55.71&68.20&63.22&66.17&63.73\\
            &ChatGPT (SC) &74.40&72.75&55.50&54.70&68.90&63.49&66.27&63.65 \\
            \midrule
            \multirow{4}{*}{\textbf{LLM API (10-shot)}} &GPT 4 &70.50&70.49&57.30&56.88&67.20&62.91&65.00&63.43 \\
            &GPT 3.5 &73.10&71.74&57.20&57.07&66.90&63.79&65.73&64.20 \\
            &ChatGPT &76.10&74.60&58.60&58.51&68.90&60.51&67.87&64.54\\
            &ChatGPT (SC) &76.60&75.07&59.10&59.04&68.80&59.56&68.17&64.55 \\
            \midrule
            \multirow{5}{*}{\textbf{Alpaca (10-shot)}}
            &MPT (7B) &43.42&34.71&48.72&37.94&65.33&43.72&52.49&38.79 \\
            &Falcon (7B) &60.68&55.07&56.35&56.15&63.92&45.17&60.32&52.13 \\
            &Mistral (7B) &76.08&74.10&59.20&58.66&67.66&60.69&67.65&64.49 \\
		  &Llama2 (7B) &61.96&61.94&55.53&53.19&69.71&60.24&62.40&58.46 \\
            &Llama2 (13B) &75.28&72.31&58.97&58.92&66.93&61.73&67.06&64.32 \\
            \midrule
            \multirow{5}{*}{\textbf{Direct Injection}}
            &MPT (7B) &63.87&63.23&53.71&52.37&51.85&51.70&56.47&55.77 \\
            &Falcon (7B) &63.48&58.54&55.27&55.16&51.21&51.14&56.66&54.95 \\
            &Mistral (7B) &74.90&74.62&59.39&59.11&59.95&59.27&64.74&64.33 \\
		  &Llama2 (7B) &67.24&66.34&56.66&55.72&55.11&55.11&59.67&59.05 \\
            &Llama2 (13B) &75.04&74.09&60.04&59.91&59.26&58.44&64.78&64.15  \\
            \midrule
            \multirow{5}{*}{\textbf{AbsInstruct}}
            &MPT (7B) &71.34&70.89&58.63&58.63&67.52&65.16&65.83&64.89 \\
            &Falcon (7B) &66.92&66.45&57.06&56.11&69.03&64.15&64.33&62.24 \\
            &Mistral (7B) &\underline{80.59}&\underline{79.85}&\textbf{60.80}&\textbf{60.74}&70.96&66.54&\underline{70.78}&\underline{69.04} \\
		  &Llama2 (7B) &77.07&75.81&59.44&59.07&\textbf{72.72}&\textbf{68.00}&69.74&67.63 \\
            &Llama2 (13B) &\textbf{81.13}&\textbf{80.35}&\underline{60.58}&\underline{60.58}&\underline{71.92}&\underline{67.24}&\textbf{71.21}&\textbf{69.39} \\
		\bottomrule
	\end{tabular}
	\caption{Performance of \papertitle and baselines on the test set of \abspyramid. We abbreviate Accuracy and Macro F1-score to \textbf{Acc} and \textbf{Ma-F1}, respectively. We bold the best score and underline the second-best score. See \cref{app:validation_results} for the performance on the validation set.}
    \label{tab:main_eval_test}
\end{table*}

\section{Experiment}
We conduct extensive experiments and compare our framework with various baselines.

\subsection{Dataset and Evaluation Metric}
We study LLMs' abstraction ability on \abspyramid, a large-scale dataset of abstraction knowledge with statistics in \cref{app:data_stat}. Our framework and baselines build examples based on five-element tuples from its training set. Meanwhile, the general-purpose instruction dataset we use is Alpaca~\cite{taori2023alpaca}, which contains 52K instructions generated with the \selfinstruct framework~\cite{wang2023self}. We mix instructions for abstraction with those general-purpose ones to fine-tune LLMs in the following experiments.

We calculate Accuracy and Macro F1-score for metrics between predicted and ground truth labels to evaluate all models' abstraction ability.

\subsection{Baseline Methods}
We compare our framework to three baselines and provide implementation details in \cref{app:implement_detail}, including learning rates, example numbers, API specifics, prompts for baselines, etc.

\paragraph{API-based LLM:} We evaluate a series of closed-source LLMs under the zero-shot and few-shot (10-shot) settings, covering GPT3.5~\cite{ouyang2022training}, ChatGPT~\cite{openai2023chatgpt}, and GPT4~\cite{achiam2023gpt4}. In addition, we test ChatGPT with the self-consistency decoding strategy~\cite{wang2022self}.

\paragraph{Alpaca LLM:} An intuitive method is to align open-source LLMs and test their abstraction ability with in-context learning. Here, we choose to tune LLMs with Alpaca~\cite{taori2023alpaca}, including models of MPT (7B)~\cite{MosaicML2023Introducing}, Falcon (7B)~\cite{penedo2023refinedweb}, Mistral (7B)~\cite{jiang2023mistral}, Llama2 (7B, 13B)~\cite{touvron2023llama}. For inference, we test models with ten exemplars randomly sampled from \abspyramid.

\paragraph{Direct Injection:} This baseline randomly samples tuples from \abspyramid and builds examples with the vanilla prompts (in \cref{app:vanilla_prompt}), where responses are solely ``Yes'' or ``No.'' Then, we mix abstraction examples with Alpaca for alignment. Similarly, the LLMs we tested are MPT (7B), Falcon (7B), Mistral (7B), and Llama2 (7B, 13B).

\begin{table}[t]
\setlength{\tabcolsep}{4.4pt}
    \small
	\centering
	\begin{tabular}{l|cccc|l}
	\toprule
        \multirow{1}{*}{\textbf{Models}}&\multicolumn{1}{c}{\textbf{Noun}} &\multicolumn{1}{c}{\textbf{Verb}}&\multicolumn{1}{c}{\textbf{Event}}&\multicolumn{1}{c}{\textbf{All}}&\multicolumn{1}{c}{\textbf{$\Delta$\textsubscript{All}}}\\
            \midrule
            Llama2 (7B) &\textbf{75.81}&\textbf{59.07}&\textbf{68.00}&\textbf{67.63}&\multicolumn{1}{c}{-} \\
            \midrule
            $\diamond$ w P-Random &69.56&58.48&66.04&64.69&$\downarrow$2.94 \\
            $\diamond$ w P-Input &69.92&58.43&66.34&64.90&$\downarrow$2.73\\
            $\diamond$ w/o Q Filter &65.06&56.90&62.70&61.55&$\downarrow$6.08 \\
            $\diamond$ w/o P\&Q Filter &65.79&57.27&54.52&59.19&$\downarrow$8.44 \\
            \midrule
            $\diamond$ w/o E Trace &69.98&58.25&66.27&64.84&$\downarrow$2.79 \\
            $\diamond$ w/o All Parts &66.34&55.72&55.11&59.05&$\downarrow$8.58\\
		\bottomrule
	\end{tabular}
	\caption{Ablation study for \papertitle. Macro F1-scores are exhibited, and \textbf{$\Delta$\textsubscript{All}} indicates score changes. See \cref{app:full_ablation_study} for results of all models.}
    \label{tab:part_ablation}
\end{table}

\section{Main Evaluation}
We present the results of each entailment relation and the average on the test set of \abspyramid in \cref{tab:main_eval_test}. In general, our framework \papertitle can unlock stronger abstraction ability from LLMs, exceeding the performance of all baselines by a large margin. For example, Mistral (7B) tuned with our framework correctly classifies 70.78\% of test examples, increasing by 6.04\% compared to the ``Direct Injection'' baseline. Meanwhile, Llama2 (13B), tuned with our framework, outperforms all the API-based LLMs, even GPT4.

Our results unequivocally demonstrate that the ``Direct Injection'' baseline possesses limited efficacy in eliciting abstraction knowledge. For example, Falcon (7B) only achieves performance slightly higher than a random guess. Similarly, we observe that LLMs tuned with Alpaca only capture limited generalization ability in abstraction detection, even with ten exemplars. For instance, Falcon (7B) only achieves a Macro F1-score of 52.13\%, lagging behind our framework by about 10 points.

\begin{figure}[t]
    \centering
    \includegraphics[width=\columnwidth]{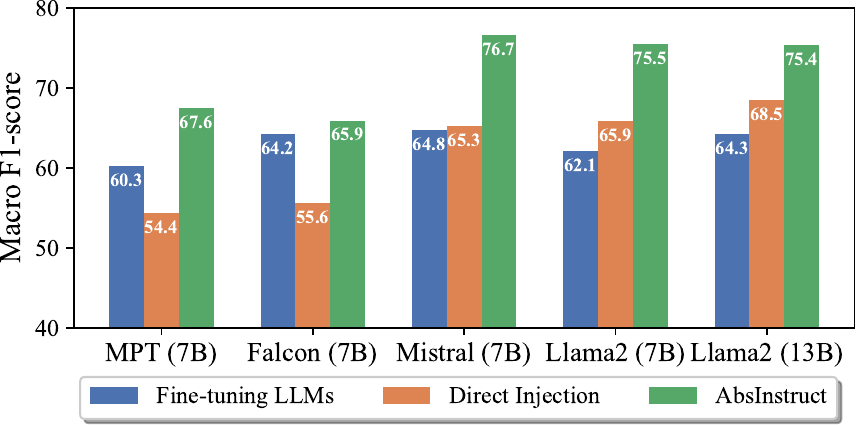}
    \caption{The out-of-domain performance on the abstractATOMIC dataset. We provide results across all metrics in \cref{app:full_ood_results}.}
    \label{fig:abstract_atomic}
\end{figure}

\subsection{Ablation Study}
To better understand how to unlock abstraction ability, we conduct several ablation experiments to show the effectiveness of explanation traces, quality filters, and plausibility estimators. The results of ablation studies are presented in \cref{tab:part_ablation}.

\paragraph{Plausibility Estimator:} We conduct two experiments to verify the efficacy of the plausibility estimator. First, we remove the estimator and randomly select examples ($\diamond$ w P-Random). From the results in \cref{tab:part_ablation}, we can find noticeable performance declines, verifying the plausibility estimator's efficacy. Moreover, we consider another way to measure plausibilities instead of normalized conditional probabilities of explanation traces. Here, we compute the normalized probabilities of example input ($\diamond$ w P-Input). As the performance consistently drops, we can see that explanation traces play a pivotal role in selecting plausible examples.

\paragraph{Quality Filters:} We also conduct ablation studies for quality filters, including the prediction and keyword filters. Results ($\diamond$ w/o Q Filter) show that LLMs' performance deteriorates drastically after we remove these filters. Then, we further remove the plausibility estimator besides quality filters ($\diamond$ w/o P\&Q Filter). The results, like the decline of 8.44\% on average, again show the efficacy of our filters and the plausibility estimator. Meanwhile, we analyze the role of the diversity filter in \cref{app:diversity_filter_study}, where we find that our framework can collect highly diverse examples and explanation traces, even without the diversity filter.

\paragraph{Explanation Traces:} First, we remove explanation traces and employ the vanilla prompt, also used by the ``Direct Injection'' baseline. The results ($\diamond$ w/o E Trace) show that LLMs cannot perform well. Further, we remove all the filters, estimator, and explanation traces ($\diamond$ w/o All Parts), where we observe greater decreases in performance. Here, Llama2 (7B) significantly drops by 8.58\% in the Macro F1-score. These findings demonstrate the utility of the explanation traces we collect.

\begin{table}[t]
\setlength{\tabcolsep}{5.2pt}
    \small
	\centering
	\begin{tabular}{l|ccll}
	\toprule
        \multirow{1}{*}{\textbf{Models}}&\multicolumn{1}{c}{\textbf{Acc}} &\multicolumn{1}{c}{\textbf{Ma-F1}}&\multicolumn{1}{c}{\textbf{$\Delta$\textsubscript{Acc}}}&\multicolumn{1}{c}{\textbf{$\Delta$\textsubscript{Ma-F1}}}\\ 
            \midrule
            \multicolumn{5}{l}{\textbf{Fine-tuned on AbsPyramid}} \\
            \midrule
            Mistral (7B) &79.32&72.66&\multicolumn{1}{c}{-}&\multicolumn{1}{c}{-}  \\
		  Llama2 (7B) &78.69&71.07&\multicolumn{1}{c}{-}&\multicolumn{1}{c}{-} \\
            Llama2 (13B) &82.11&71.25&\multicolumn{1}{c}{-}&\multicolumn{1}{c}{-} \\
            \midrule
            \multicolumn{1}{l}{\textbf{Direct Injection}}\\
            \midrule
            Mistral (7B) &85.34&74.55&$\uparrow$6.02&$\uparrow$1.89 \\
		  Llama2 (7B)&84.29&74.00&$\uparrow$5.60&$\uparrow$2.93 \\
            Llama2 (13B) &85.51&76.27&$\uparrow$3.40&$\uparrow$5.02 \\
            \midrule
            \multicolumn{1}{l}{\textbf{\papertitle}}\\
            \midrule
            Mistral (7B)&86.61&77.80&$\uparrow$\textbf{7.29}&$\uparrow$5.14\\
            Llama2 (7B)&84.31&78.76&$\uparrow$5.62&$\uparrow$7.69\\
            Llama2 (13B)&\textbf{87.11}&\textbf{79.89}&$\uparrow$5.00&$\uparrow$\textbf{8.64}\\
		\bottomrule
	\end{tabular}
	\caption{The performance on the Levy/Holt dataset. \textbf{$\Delta$\textsubscript{Acc}} and \textbf{$\Delta$\textsubscript{Ma-F1}} mean improvements compared to LLMs fine-tuned on \abspyramid. We show results of all LLMs in \cref{app:full_ood_results}}
    \label{tab:levy_holt}
\end{table}

\subsection{Out-of-Domain Evaluation}
This section studies if our framework can generalize to other tasks requiring abstraction knowledge. We conduct experiments on two out-of-domain datasets: AbstractATOMIC~\cite{he2022acquiring} and Levy/Holt dataset~\cite{levy2016annotating, holt2018probabilistic}, with statistics in \cref{app:ood_stat}.

\paragraph{AbstractATOMIC:} First, we test our framework on the AbstractATOMIC dataset and treat ``Direct Injection'' as a baseline. We also fine-tune LLMs on \abspyramid to test their transferring ability on AbstractATOMIC. As depicted in \cref{fig:abstract_atomic}, our framework can equip LLMs with broader generalization abilities. Particularly, Mistral (7B) attains a Macro F1-score of 76.7\%, which is substantially higher than ``Direct Injection.'' Also, Llama2 (7B) exhibits improvements of over 10 points compared to its fine-tuned counterpart, which demonstrates our work's essence of eliciting abstraction ability instead of fitting a specific dataset.

\paragraph{Levy/Holt Dataset:} This dataset is primarily used to evaluate verb entailment graphs. We test the performance of models tuned with our framework and take the same baselines as AbstractATOMIC. As shown in \cref{tab:levy_holt}, our framework performs better on the Levy/Holt dataset than the ``Direct Injection'' baseline. More generally, instruction-tuning methods can obtain better generalization than fine-tuning on \abspyramid, given that instruction-tuning only needs a tiny fraction of training data. With our framework, the Macro F1-score of Llama2 (13B) improves considerably by 8.64\% compared to the fine-tuned one.

\subsection{Discussion of Explanation Trace}
\begin{figure}[t]
    \centering
    \includegraphics[width=\columnwidth]{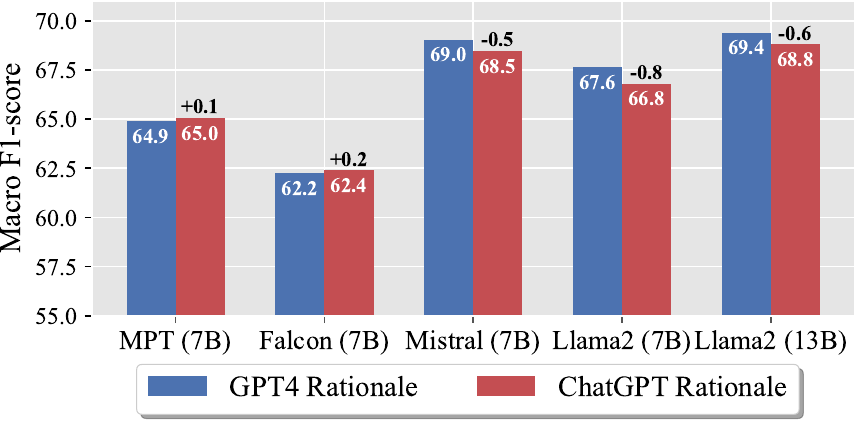}
    \caption{Macro F1-scores of our framework with ChatGPT as the source of explanation traces. We also provide performance changes. See full results across all metrics in \cref{app:full_chatgpt_rationale}.}
    \label{fig:chatgpt_rationale_score}
\end{figure}

In previous sections, we collect explanation traces by prompting GPT4. Here, we evaluate our framework with explanation traces from a less advanced model, ChatGPT, to gain a deeper insight into the robustness. We plot and compare the Macro F1-scores in \cref{fig:chatgpt_rationale_score}. The outcomes suggest that our framework maintains its strong performance with some fluctuations below 1 point. In particular, the score of Falcon (7B) improves by only 0.2\% while Llama2 (13B) declines by only 0.6\%. 

\subsection{Task Instruction Following}
Prior experiments demonstrate the effectiveness of our framework in abstraction ability. Additionally, we also evaluate the ability of LLMs to follow general-purpose instructions for NLP tasks. Here, we choose the test set of SuperNI~\cite{wang2022super}, consisting of 119 tasks with 100 examples in each task. Following previous works~\cite{wang2023self,xu2023wizardlm}, we evaluate LLMs by calculating ROUGE-L~\cite{lin2004rouge}, BLEU-1/2~\cite{papineni2002bleu}, and Meteor~\cite{banerjee2005meteor}. For baselines, we train LLMs on the instruction dataset Alpaca~\cite{taori2023alpaca}, as it is also used in our framework. The results in \cref{tab:superni_performance} show that LLMs tuned with our framework can attain comparable scores to those fine-tuned on Alpaca. For instance, MPT (7B) obtains a slightly higher Rouge-L while Llama2 (7B) drops by only 0.09\%. These findings manifest that injecting a few instructions for abstraction knowledge does not sacrifice the general ability of instruction following.

\begin{table}[t]
    \small
    \setlength{\tabcolsep}{3.1pt}
	\centering
	\begin{tabular}{l|ccccc}
	    \toprule
		\textbf{Models}&\textbf{R-L}&\textbf{B-1}&\textbf{B-2}&\textbf{Meteor}&\textbf{$\Delta$\textsubscript{R-L}}\\
            \midrule
            \multicolumn{6}{l}{\textbf{\textsc{Alpaca}}} \\
            \midrule
            MPT (7B) &41.20&26.44&14.37&26.20&- \\
		Falcon (7B) &39.38&24.21&12.88&25.19&- \\
            Mistral (7B) &50.47&\textbf{44.66}&\textbf{26.83}&31.06&- \\
		  Llama2 (7B) &43.70&28.21&15.19&27.37&- \\
            Llama2 (13B) &48.30&30.70&17.32&30.39&- \\
            \midrule
            \midrule
            \multicolumn{6}{l}{\textbf{\textsc{AbsInstruct}}} \\
            \midrule
            MPT (7B) &43.43&26.40&14.40&27.71&$\uparrow$\textbf{1.51} \\
		Falcon (7B) &39.76&26.52&14.38&25.66&$\uparrow$0.47\\
            Mistral (7B) &\textbf{51.22}&42.58&24.99&\textbf{31.88}&$\uparrow$0.82 \\
		  Llama2 (7B) &43.35&26.80&14.29&27.28&$\downarrow$0.09 \\
            Llama2 (13B) &49.19&31.71&17.66&31.25&$\uparrow$0.86 \\
		\bottomrule
	\end{tabular}
	\caption{Performance on the test set of SuperNI. \textbf{R-L} and \textbf{B-1/2} denote ROUGE-L and BLEU-1/2. \textbf{$\Delta$\textsubscript{R-L}} means the performance changes compared to Alpaca.}
    \label{tab:superni_performance}
\end{table}

Meanwhile, previous works~\cite{ouyang2022training, zhao2023survey} suggest a disparity between NLP tasks and human requests. Thus, we also conduct a human evaluation on expert-curated instructions~\cite{wang2023self} to better understand the alignment with human values. The evaluation setups and results are shown in \cref{app:instruct_human_evaluation}, which again manifests that our framework can preserve LLMs' general capabilities.


\section{Conclusion}
Abstraction knowledge is a critical kind of knowledge in human intelligence, as shown in previous cognitive studies~\cite{minsky1980k}. Meanwhile, recent research~\cite{zheng2023take, gao2024efficient} shows that using LLMs' abstraction ability can better solve general NLP tasks, including STEM questions~\cite{hendrycks2020measuring, miao2020diverse}, Knowledge QA~\cite{yang2018hotpotqa, kwiatkowski2019natural, joshi2017triviaqa}, and Multi-Hop Reasoning~\cite{trivedi2022musique, geva2021did}. Given this evidence, we can see that abstraction is also essential for a broad range of complex language understanding and reasoning tasks. 

In this paper, we propose \papertitle, which is the first attempt to elicit stronger abstraction abilities from pre-trained LLMs. Our framework builds instructions for abstraction detection with explanation traces and a plausibility estimator. Then, these abstraction instructions are combined with general-domain ones from Alpaca. In the experiments, we compare our framework with a lot of strong baselines to demonstrate the framework's effectiveness. We also provide comprehensive ablation studies of our framework and show its effectiveness on two out-of-domain datasets. What's more, evaluations on instruction datasets also show that our framework can improve the abstraction ability of LLMs without sacrificing LLMs' general instruction following ability.

For future work, we can study how to equip LLMs with more abstraction knowledge during pre-training. More importantly, we also leave the study of using this enhanced abstraction knowledge in downstream tasks as future work.

\section*{Limitations}
Prior research~\cite{zhou2023lima} indicates that LLMs primarily acquire their knowledge during the pre-training phase, while the alignment phase only teaches LLMs about the specific subdistribution of interactions with users. In this work, we mainly focus on the alignment phase, while it remains unclear what abstraction knowledge is captured by LLMs during pre-training. Following previous works of knowledge probing~\cite{hou2023towards,sun2023head}, future research can probe recent LLMs, like Llama2, to better understand this question and explore how to equip LLMs with more abstraction knowledge during pre-training.

Meanwhile, instruction tuning only elicits the existing knowledge of pre-trained LLMs. We leave for future works about equipping LLMs with new abstraction knowledge through other techniques, like knowledge editing~\cite{wang2023zekun,zhang2024comprehensive,hase2023does}, retrieval augmented generation~\cite{lewis2020retrieval,gao2023retrieval,wu2024scimmir}, event-centric knowledge~\cite{wang2023cola,wang2022subeventwriter, gao2023chatgpt, do2024constraintchecker}, intention detection~\cite{wu2023new}, and knowledge population~\cite{shen2023dense}. Meanwhile, we can extend our abstraction knowledge to multimodal, like exploring knowledge from given images~\cite{shen2024vcd, cui2024open}. 

\section*{Ethics Statement}
We evaluate the abstraction ability on \abspyramid~\cite{wang2023abspyramid}, which is a free and open-source dataset. The out-of-domain (OOD) datasets, namely AbstractATOMIC~\cite{he2022acquiring} and Levy/Holt~\cite{levy2016annotating,holt2018probabilistic}, are also freely available and open-source. The instruction datasets SuperNI~\cite{wang2022super} and \selfinstruct~\cite{wang2023self} are released under the Apache-2.0 License. Meanwhile, the Alpaca~\cite{taori2023alpaca} dataset is released under the CC BY NC 4.0 License.

Human evaluations are performed by three expert annotators with at least one year of expertise in NLP to ensure quality. The annotation works are compensated at the hourly rate of 7.6 USD, higher than the local minimum wage.

\section*{Acknowledgements}
The authors of this paper were supported by the NSFC Fund (U20B2053) from the NSFC of China, the RIF (R6020-19 and R6021-20), and the GRF (16211520 and 16205322) from RGC of Hong Kong. This paper was also supported by the Tencent AI Lab Rhino-bird Focused Research Program. We also thank the support from NVIDIA AI Technology Center (NVAITC) and the UGC Research Matching Grants (RMGS20EG01-D, RMGS20CR11, RMGS20CR12, RMGS20EG19, RMGS20EG21, RMGS23CR05, RMGS23EG08).

\newpage
\bibliography{anthology,custom}
\bibliographystyle{acl_natbib}

\appendix

\newpage
\section{\papertitle Prompts}
\label{app:method}
This appendix lists the concrete prompts we use in our framework. First, we provide the prompts of building instructions, input, and responses. Then, we show the concrete prompts we use to collect word meanings from GPT4.

\subsection{Prompts for Instructions and Examples}
\label{app:abs_instruct_prompt}
We manually collect the prompt templates of instructions, input, and responses used in our framework, shown in \cref{tab:abs_instruct_prompt}. Models are given five-element tuples on the \abspyramid dataset: (\textitbf{head event}, \textitbf{entailment relation}, \textitbf{tail event}, \textitbf{instance}, \textitbf{concept}). 

In our prompt templates in \cref{tab:abs_instruct_prompt}, there are three placeholders \textbf{[head]}, \textbf{[cpt]}, and \textbf{[ins]} for head events, concepts, and instances. Specifically, \textbf{[ins]} is the same as \textbf{[head]} for \eventrelation. Meanwhile, we indicate the entailment relations implicitly by using different instructions for different relations. For example, for \nounrelation, the instruction contains ``Identify the hypernym of a specific \textbf{noun}.'' Note that the tail event can be built by replacing the instance with the concept in the head event. In conclusion, our prompt does not lose any information provided by five-element tuples. 

\subsection{Prompts for Word Meanings}
\label{app:word_meaning_prompt}
To build explanation traces, we also prompt GPT4 to collect the meanings of instances and concepts in a zero-shot manner. Here, we ask GPT4 to provide meanings of given words and then detect whether the given concept is valid. The prompt is shown in \cref{tab:zero_shot_gpt4_prompt}. We collect the meanings of instances and concepts in the first and second steps separately. Then, we concatenate them to build explanation traces.

\begin{table}[t!]
\centering
\small
\begin{subtable}[t]{0.96\columnwidth}
\centering
\begin{tabular}{p{0.96\columnwidth}}
\toprule
\textbf{\nounrelation Instruction}: Hypernyms are words with a broad meaning, which more specific words fall under. Identify the hypernym of a specific noun through the following two steps: Step 1: Let's think about meanings of those words. Step 2: Provide a ``Yes'' or ``No'' response.
\\ \midrule
\textbf{\verbrelation Instruction}: Hypernyms are words with a broad meaning, which more specific words fall under. Identify the hypernym of a specific verb through the following two steps: Step 1: Let's think about meanings of those words. Step 2: Provide a ``Yes'' or ``No'' response.
\\ \midrule
\textbf{\eventrelation Instruction}: Identify abstract descriptions of specific sentences through the following two steps: Step 1: Let's think about meanings of the sentence and the abstract description. Step 2: Provide a ``Yes'' or ``No'' response. \\
\bottomrule
\end{tabular}
\caption{Instructions used by our framework.}
\end{subtable}

\begin{subtable}[t]{0.96\columnwidth}
\centering
\begin{tabular}{p{0.96\columnwidth}}
\toprule
\textbf{\nounrelation Input}: In the sentence \textbf{[head]}, does the meaning of \textbf{[cpt]} encompass \textbf{[ins]}?
\\ \midrule
\textbf{\verbrelation Input}: In the sentence \textbf{[head]}, does the meaning of \textbf{[cpt]} encompass \textbf{[ins]}?
\\ \midrule
\textbf{\eventrelation Input}: Can we consider \textbf{[cpt]} as an abstract description of the sentence \textbf{[head]}?
\\
\bottomrule
\end{tabular}
\caption{Input templates used by our framework.}
\end{subtable}

\begin{subtable}[t]{0.96\columnwidth}
\centering
\begin{tabular}{p{0.96\columnwidth}}
\toprule
\textbf{\nounrelation, \verbrelation, and \eventrelation Response}
\newline
\textbf{Positive Label:} Step1: \textbf{<ins mean>}. Meanwhile, \textbf{<cpt mean>}. Step2: Yes, the meaning of \textbf{[cpt]} encompasses \textbf{[ins]}.
\newline
\textbf{Negative Label:} Step1: \textbf{<ins mean>}. Meanwhile, \textbf{<cpt mean>}. Step2: No, the meaning of \textbf{[cpt]} does not encompass \textbf{[ins]}.
\\
\bottomrule
\end{tabular}
\caption{Response templates used by our framework.}
\end{subtable}

\caption{The concrete prompts we used in our \papertitle framework. We show the instruction, input, and response templates in each table segment. Placeholders \textbf{[head]}, \textbf{[cpt]}, and \textbf{[ins]} will be replaced with real head events, concepts, and instances. Also, \textbf{<ins mean>} and \textbf{<cpt mean>} will be replaced with the meanings of real instances and concepts.}
\label{tab:abs_instruct_prompt}
\end{table}

\begin{table}[t!]
\centering
\small
\begin{tabular}{p{0.96\columnwidth}}
\toprule
\textbf{\nounrelation}: Identify the hypernym of a specific noun. Hypernyms are words with a broad meaning, which more specific words fall under. In the sentence \textbf{[head]}, does the meaning of the new word \textbf{[cpt]} encompass the original word \textbf{[ins]}? 
\newline Step 1: Let's think about the meaning of the original word.
\newline Step 2: Let's think about the meaning of the new word.
\newline Step 3: Provide a ``Yes'' or ``No'' response without other words.
\\ \midrule
\textbf{\verbrelation}: Identify the hypernym of a specific verb. Hypernyms are words with a broad meaning, which more specific words fall under. In the sentence \textbf{[head]}, does the meaning of the new word \textbf{[cpt]} encompass the original word \textbf{[ins]}? 
\newline Step 1: Let's think about the meaning of the original word. 
\newline Step 2: Let's think about the meaning of the new word. 
\newline Step 3: Provide a ``Yes'' or ``No'' response without other words.
\\ \midrule
\textbf{\eventrelation}: Identify abstract descriptions of specific sentences. Can we consider \textbf{[cpt]} as an abstract description of the sentence \textbf{[head]}? 
\newline Step 1: Let's think about the meaning of the sentence. 
\newline Step 2: Let's think about the meaning of the abstract description. 
\newline Step 3: Provide a ``Yes'' or ``No'' response without other words.
\\
\bottomrule
\end{tabular}

\caption{The zero-shot prompts we used for collecting meanings of instances and concepts. Placeholders \textbf{[head]}, \textbf{[cpt]}, and \textbf{[ins]} will be replaced with real head events, concepts, and instances. We collect the meanings of instances and concepts in the first and second steps separately. Then, we concatenate them to build explanation traces.}
\label{tab:zero_shot_gpt4_prompt}
\end{table}

\section{Human Annotation}
We conduct a few human evaluations in our study, including the Accuracy of GlossBERT~\cite{huang2019glossbert}, the quality of examples collected by our framework, and the ability of our framework to follow human instructions. In this appendix, we discuss the details of annotation and agreement between annotators. 

All annotation tasks are performed by three post-graduate NLP researchers with at least one year of expertise in NLP. They understand our annotation tasks clearly and can serve as experts. Two annotators are authors of the paper, and the third is another NLP researcher within the same institution for a more objective perspective. The authors’ involvement in the annotation process is part of their academic responsibilities, and no additional compensation is provided. The third annotator is compensated at the hourly rate of 7.6 USD, higher than the local minimum wage.

\paragraph{GlossBERT Accuracy:} \label{app:wsd_annotation}
We sample 500 examples from \abspyramid and run GlossBERT to disambiguate the given noun or verb. Three experts are asked to evaluate whether the disambiguation results are right, yielding 1500 ratings in total. The IAA score is 78.8\% calculated using pairwise agreement proportion, and the Fleiss's $\kappa$~\cite{fleiss1971measuring} is 0.57.

\paragraph{Quality of Collected Examples:} 
\label{app:quality_collected_example}
We sampled 150 explanation traces collected by our framework \papertitle. Similarly, three experts are asked to label two aspects: the correctness of explanations for given instances and concepts. This leads to 900 total ratings (150 examples $\times$ 2 aspects $\times$ 3 annotators). The Fleiss's $\kappa$~\cite{fleiss1971measuring} is 0.62.

\paragraph{Human Instruction Following:} 
\label{app:human_instruction_following}
There are 252 instructions in the test set of \selfinstruct~\cite{wang2023self}, which are curated manually by experts. Here, the expert annotators are asked to annotate which response is preferred between our framework and the Alapca baseline. This leads to 756 ratings for each model. The IAA score is 80.95\%
calculated using pairwise agreement proportion, and the Fleiss’s $\kappa$~\cite{fleiss1971measuring} is 0.71.

\section{Implementation Details}
\label{app:implement_detail}
We access open-source language models using Transformers~\cite{wolf2020transformers} and fine-tune them on 8 NVIDIA A100 (80G) GPUs. We fine-tune 7B and 13B LLMs with LoRA~\cite{hu2021lora} and load them with BF16. For LoRA, we only add new parameters to attention layers with the rank and $\alpha$ equal to 512 and 1024. The best checkpoint is selected according to the sum of all metrics on the validation set. The batch size and training epoch are 128 and 3, respectively. We grid search learning rates of 5e-6, 1e-5, 2e-5, 3e-5 and 5e-5. 

We collect rationales for about 2,000 examples for each entailment relation and keep 200 examples with the highest LLM-intrinsic plausibility after filtering. For a fair comparison, the ``Direct Injection'' baseline also incorporates 200 examples for each entailment relation. We discuss choices of example numbers and show that 200 is proper in \cref{app:instruct_number}. For API-based LLMs, We access ChatGPT, GPT4, and GPT3.5 via OpenAI API\footnote{https://platform.openai.com/docs/api-reference}. The specific versions are \texttt{gpt-3.5-turbo-0613}, \texttt{gpt-4-1106-preview}, and \texttt{gpt-3.5-turbo-instruct-0914}. They are evaluated on one thousand examples that we randomly sampled from the test set of each entailment relation due to the trade-off between API expenses and our evaluation's precision. For self-consistency, we sample 5 responses independently for each example and take the majority vote.

\begin{table}[t!]
\centering
\small
\begin{subtable}[t]{0.96\columnwidth}
\centering
\begin{tabular}{p{0.96\columnwidth}}
\toprule
Below is an instruction that describes a task, paired with an input that provides further context. Write a response that appropriately completes the request.
\newline
\newline
\#\#\# Instruction:
\newline
\textbf{\{instruction\}}
\newline
\newline
\#\#\# Input:
\newline
\textbf{\{input\}}
\newline
\newline
\#\#\# Response: \\
\bottomrule
	\end{tabular}
\caption{Template for examples with a non-empty input field.}
\end{subtable}

\begin{subtable}[t]{0.96\columnwidth}
\centering
\begin{tabular}{p{0.96\columnwidth}}
\toprule
Below is an instruction that describes a task. Write a response that appropriately completes the request.
\newline
\newline
\#\#\# Instruction:
\newline
\textbf{\{instruction\}}
\newline
\newline
\#\#\# Response: \\
\bottomrule
\end{tabular}
\caption{Template for examples with an empty input field.}
\end{subtable}
\caption{The prompt templates we used to concatenate instructions and example input. We show two templates since the input is optional. Placeholders \textbf{\{instruction\}} and \textbf{\{input\}} will be replaced with real instructions and example input.}
\label{tab:concatenation_prompt}
\end{table}

\begin{table}[t!]
\centering
\small
\begin{subtable}[t]{0.96\columnwidth}
\centering
\begin{tabular}{p{0.96\columnwidth}}
\toprule
\textbf{\nounrelation Instruction}: Identify the hypernym of a specific noun and provide a ``Yes'' or ``No'' response. Hypernyms are words with a broad meaning, which more specific words fall under. 
\\ \midrule
\textbf{\verbrelation Instruction}: Identify the hypernym of a specific verb and provide a ``Yes'' or ``No'' response. Hypernyms are words with a broad meaning, which more specific words fall under.
\\ \midrule
\textbf{\eventrelation Instruction}: Identify abstract descriptions of specific sentences, and provide a ``Yes'' or ``No'' response. \\
\bottomrule
\end{tabular}
\caption{Instructions of the vanilla prompt.}
\end{subtable}

\begin{subtable}[t]{0.96\columnwidth}
\centering
\begin{tabular}{p{0.96\columnwidth}}
\toprule
\textbf{\nounrelation Input}: In the sentence \textbf{[head]}, does the meaning of \textbf{[cpt]} encompass \textbf{[ins]}?
\\ \midrule
\textbf{\verbrelation Input}: In the sentence \textbf{[head]}, does the meaning of \textbf{[cpt]} encompass \textbf{[ins]}?
\\ \midrule
\textbf{\eventrelation Input}: Can we consider \textbf{[cpt]} as an abstract description of the sentence \textbf{[head]}?
\\
\bottomrule
\end{tabular}
\caption{Input templates of the vanilla prompt.}
\end{subtable}

\begin{subtable}[t]{0.96\columnwidth}
\centering
\begin{tabular}{p{0.96\columnwidth}}
\toprule
\textbf{\nounrelation, \verbrelation, and \eventrelation Response}
\newline
\textbf{Positive Label:} Yes.
\newline
\textbf{Negative Label:} No.
\\
\bottomrule
\end{tabular}
\caption{Responses of the vanilla prompt.}
\end{subtable}

\caption{The vanilla prompt we used in the ``Direct Injection'' baseline. We show the instruction, input, and response templates in each table segment. Placeholders \textbf{[head]}, \textbf{[cpt]}, and \textbf{[ins]} will be replaced with real head events, concepts, and instances.}
\label{tab:vanilla_prompt}
\end{table}

\begin{table}[t!]
\small
\centering
\begin{tabular}{p{0.96\columnwidth}}
	\toprule
	\textbf{\nounrelation:} Identify the hypernym of a specific noun and provide a ``Yes'' or ``No'' response. Hypernyms are words with a broad meaning, which more specific words fall under. In the sentence \textbf{[head]}, does the meaning of \textbf{[cpt]} encompass \textbf{[ins]}? \\
    \midrule
    \textbf{\verbrelation: } Identify the hypernym of a specific verb and provide a ``Yes'' or ``No'' response.
    Hypernyms are words with a broad meaning, which more specific words fall under. In the sentence \textbf{[head]}, does the meaning of \textbf{[cpt]} encompass \textbf{[ins]}? \\
    \midrule
    \textbf{\eventrelation: } Identify abstract descriptions of specific sentences, and provide a ``Yes'' or ``No'' response. Can we consider \textbf{[cpt]} as an abstract description of the sentence \textbf{[head]}? \\
    \bottomrule
\end{tabular}
\caption{The zero-shot prompt we used in the ``API-based LLM'' baseline. Placeholders \textbf{[head]}, \textbf{[ins]}, and \textbf{[cpt]} will be replaced with real head events, instances, and concepts.}
\label{tab:zero_shot_api_based_prompt}
\end{table}

\begin{table}[t!]
\small
\centering
\begin{tabular}{p{0.96\columnwidth}}
	\toprule
	\multicolumn{1}{c}{\textbf{\nounrelation:}}
 \\ \midrule
 \textbf{Instruction:} You need to decide whether a hypernym of a specific noun is valid or not. Hypernyms are words with a broad meaning, which more specific words fall under.
 \\ \midrule
  \textbf{Exemplars and test example:} 
  \newline
  1. In the sentence \textbf{[head]\textsuperscript{(1)}}, is \textbf{[cpt]\textsuperscript{(1)}} a hypernym of \textbf{[ins]\textsuperscript{(1)}}? Yes. (No.)
  \newline
  2. In the sentence \textbf{[head]\textsuperscript{(2)}}, is \textbf{[cpt]\textsuperscript{(2)}} a hypernym of \textbf{[ins]\textsuperscript{(2)}}? Yes. (No.)
  \newline
  \ldots
  \newline
  11. In the sentence \textbf{[head]\textsuperscript{(11)}}, is \textbf{[cpt]\textsuperscript{(11)}} a hypernym of \textbf{[ins]\textsuperscript{(11)}}?
 \\ \bottomrule \toprule
    \multicolumn{1}{c}{\textbf{\verbrelation}}
\\ \midrule 
\textbf{Instruction:} You need to decide whether a hypernym of a specific verb is valid or not. Hypernyms are words with a broad meaning, which more specific words fall under.
\\ \midrule
1. In the sentence \textbf{[head]\textsuperscript{(1)}}, is \textbf{[cpt]\textsuperscript{(1)}} a hypernym of \textbf{[ins]\textsuperscript{(1)}}? Yes. (No.)
  \newline
  2. In the sentence \textbf{[head]\textsuperscript{(2)}}, is \textbf{[cpt]\textsuperscript{(2)}} a hypernym of \textbf{[ins]\textsuperscript{(2)}}? Yes. (No.)
  \newline
  \ldots
  \newline
  11. In the sentence \textbf{[head]\textsuperscript{(11)}}, is \textbf{[cpt]\textsuperscript{(11)}} a hypernym of \textbf{[ins]\textsuperscript{(11)}}?
\\ \bottomrule \toprule
    \multicolumn{1}{c}{\textbf{\eventrelation}} 
\\ \midrule
\textbf{Instructions:} You need to decide whether an abstract description of a specific sentence is valid or not.
\\ \midrule
1. Can we consider \textbf{[cpt]\textsuperscript{(1)}} as an abstract description of the sentence \textbf{[head]\textsuperscript{(1)}}? Yes. (No.)
\newline
2. Can we consider
\textbf{[cpt]\textsuperscript{(2)}} as an abstract description of the sentence \textbf{[head]\textsuperscript{(2)}}? Yes. (No.)
\newline
\ldots
\newline
11. Can we consider
\textbf{[cpt]\textsuperscript{(11)}} as an abstract description of the sentence \textbf{[head]\textsuperscript{(11)}}?
\\
\bottomrule
\end{tabular}

\caption{The in-context learning prompt (10-shot) we used in the ``API-based LLM'' baseline. Placeholders \textbf{[head]}, \textbf{[ins]}, and \textbf{[cpt]} will be replaced with real head events, instances, and concepts.}
\label{tab:icl_api_based_prompt}
\end{table}

\subsection{Prompts for Concatenation}
\label{app:prompt_for_concatenation}
We should concatenate the instructions and input as a prompt for our framework and the instruction-tuned baselines: ``Alpaca LLM'' and ``Direct Injection.'' In our experiments, we employ the same prompt template as used by Alpaca~\cite{taori2023alpaca}, which is shown in \cref{tab:concatenation_prompt}.

\subsection{The Vanilla Prompt of the ``Alpaca LLM'' and ``Direct Injection'' Baselines}
\label{app:vanilla_prompt}
This appendix provides the vanilla prompt used by the ``Alpaca LLM'' and ``Direct Injection'' baseline to build instructions and examples for abstraction detection, as demonstrated in \cref{tab:vanilla_prompt}. In contrast to our framework, the responses of this vanilla prompt are simply ``Yes'' or ''No,'' verbalized directly from the binary labels.

\subsection{The Prompt of the ``API-based LLM'' Baseline}
\label{app:api_based_llm_prompt}
We employ the same prompt as those utilized in \abspyramid~\cite{wang2023abspyramid}, which exhibit considerable robustness when benchmarked against other prompts featured in the study of \abspyramid.
We provide zero-shot prompts used by the ``API-based LLM'' baseline in \cref{tab:zero_shot_api_based_prompt}. The in-context learning prompts used by the ``API-based LLM'' baseline are shown in \cref{tab:icl_api_based_prompt}. 

\begin{figure}[t]
    \centering
    \includegraphics[width=\columnwidth]{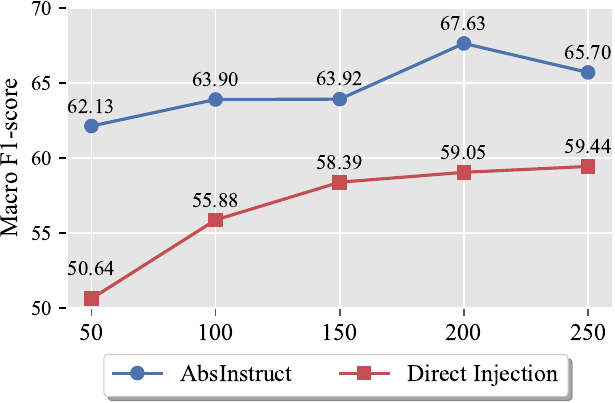}
    \caption{Macro F1-scores of different example numbers. We instruction-tune Llama2 (7B) with our \papertitle framework and the ``Direct Injection'' baseline.}
    \label{fig:k_value_study}
\end{figure}

\subsection{Discussion of Example Number $K$}
\label{app:instruct_number}
In our experiments, we collect $K=200$ examples of abstraction detection for each relation in our framework and the ``Direct Injection'' baseline. In this appendix, we study the proper values of this hyperparameter and grid search different example numbers $K$ of 50, 100, 150, 200, and 250. Here, we combine different numbers of abstraction instructions with Alpaca and instruction-tune Llama2 (7B) with our \papertitle framework and the ``Direct Injection'' baseline. We plot the performance in \cref{fig:k_value_study}. For the ``Direct Injection'' baseline, the Macro F1-score shows rapid improvements as the number of examples increases from 50 to 200. Then, the improvement stagnates (i.e., lower than 0.5 points) at the number of 250. Similarly, for our framework, we observe the first decrease when the example number is 250. For both methods, we can find that the improvements are small when the example numbers are higher than 200. Thus, we recommend choosing 200 examples for each relation in our framework and the ``Direct Injection'' baseline when considering the tradeoff between abstraction ability and general-domain ability.

\section{Supplementary Experiments}
\label{app:supplementary_experiment}
This appendix provides more supplementary experiments and analysis of the \papertitle framework.
\begin{table}[t]
    \small
    \setlength{\tabcolsep}{5pt}
    \centering
	\begin{tabular}{l|ccccc}
	    \toprule
            \textbf{\textsc{Split}}&\textbf{Noun}&\textbf{Verb}&\textbf{Event}&\textbf{All}&\textbf{Pos \%}\\
		\midrule
		\textsc{Train}& 79,034 & 47,669 & 49,988 & 176,691 & 58.86 \\
		  \textsc{Valid}& 9,874 & 5,939 & 6,237 & 22,050 &58.22 \\
            \textsc{Test}& 9,875 & 5,934 & 6,247 & 22,056 &59.02 \\
            \midrule
            Ours & 200 & 200 & 200 & 600 & 50.00 \\
		\bottomrule
	\end{tabular}
	\caption{Statistics of \abspyramid and abstraction examples collected by \papertitle. \textbf{Pos \%} denotes positive rates of each split. Our framework samples 200 examples for each relation with balanced labels.}
	\label{table:abspyramid_stat}
\end{table}

\begin{table}[t]
    \small
    \setlength{\tabcolsep}{4pt}
    \centering
	\begin{tabular}{l|cccc}
	    \toprule
        \textbf{Datasets}&\textbf{\# Total}&\textbf{\# Train}&\textbf{\# Valid}&\textbf{\# Test}\\
		\midrule
  	AbsAtomic &92235&75814&8027&8394 \\
        Levy/Holt &18407&N/A&5486&12921\\
		\bottomrule
	\end{tabular}
	\caption{Statistics of AbstractATOMIC and Levy/Holt datasets. \textbf{\# Total} is the number of all examples.}
	\label{table:ood_stat}
\end{table}

\subsection{Abstraction Data Statistics}
\label{app:data_stat}
We study LLMs' abstraction ability on \abspyramid~\cite{wang2023abspyramid}, a large-scale benchmark of abstraction knowledge comprising more than 221K examples. The dataset samples head events from ASER~\cite{zhang2020aser,zhang2022aser} and collects abstract concepts of three components of head events: nouns, verbs, and entire events. \abspyramid collects candidates of abstract concepts using WordNet~\cite{miller1995wordnet} and ChatGPT~\cite{openai2023chatgpt}, which is then manually verified. Our framework only builds 200 examples for each entailment relation based on five-element tuples from the training split. We present comprehensive statistics of \abspyramid and our examples in \ref{table:abspyramid_stat}.

\subsection{Out-of-Domain Datasets Statistics}
\label{app:ood_stat}
As we conduct experiments on two out-of-domain datasets: AbstractATOMIC~\cite{he2022acquiring} and Levy/Holt dataset~\cite{levy2016annotating, holt2018probabilistic}, we provide comprehensive statistics of these two datasets in \cref{table:ood_stat}. AbstractATOMIC samples base events from ATOMIC~\cite{sap2019atomic} in the commonsense domain and collects thousands of abstract concepts for nouns and entire events. Meanwhile, the Levy/Holt dataset is primarily used in the study of verb entailment graphs, where events are simplified as a verb with two entity types as arguments (i.e., \textit{subject} and \textit{object}).

\subsection{Validation Results on Abstraction Detection}
\label{app:validation_results}
We collect the performance of our framework \papertitle and baselines on the validation set of the \abspyramid in \cref{tab:main_eval_validation}.

\begin{table}[t]
\setlength{\tabcolsep}{4.4pt}
    \small
	\centering
	\begin{tabular}{l|cccc|l}
	\toprule
        \multirow{1}{*}{\textbf{Models}}&\multicolumn{1}{c}{\textbf{Noun}} &\multicolumn{1}{c}{\textbf{Verb}}&\multicolumn{1}{c}{\textbf{Event}}&\multicolumn{1}{c}{\textbf{All}}&\multicolumn{1}{c}{\textbf{$\Delta$\textsubscript{All}}}\\
            \midrule
		  MPT (7B) 
    &\textbf{70.89}&\textbf{58.63}&\textbf{65.16}&\textbf{64.89}&\multicolumn{1}{c}{-} \\
            \midrule
            $\diamond$ w P-Random &69.57&57.89&55.26&60.90&$\downarrow$3.99 \\
            $\diamond$ w P-Input &70.06&58.32&59.52&62.63&$\downarrow$2.26 \\
            $\diamond$ w/o Q Filter &60.24&53.02&54.40&55.89&$\downarrow$9.00 \\
            $\diamond$ w/o P\&Q Filter &58.61&48.06&32.42&46.37&$\downarrow$18.52 \\
            \midrule
            $\diamond$ w/o E Trace &65.46&54.77&63.10&61.11&$\downarrow$3.78 \\
            $\diamond$ w/o All Parts &63.23&52.37&51.70&55.77&$\downarrow$9.12 \\
            \midrule
            \midrule
            Falcon (7B) &\textbf{66.45}&\textbf{56.11}&\textbf{64.15}&\textbf{62.24}& \multicolumn{1}{c}{-} \\
            \midrule
            $\diamond$ w P-Random &61.85&55.53&62.30&59.89&$\downarrow$2.35 \\
            $\diamond$ w P-Input &61.25&53.92&58.95&58.04&$\downarrow$4.20 \\
            $\diamond$ w/o Q Filter &52.81&39.83&58.25&50.30&$\downarrow$11.94 \\
            $\diamond$ w/o P\&Q Filter &59.50&50.41&59.36&56.42&$\downarrow$5.82 \\
            \midrule
            $\diamond$ w/o E Trace &62.89&52.75&61.18&58.94&$\downarrow$3.30 \\
            $\diamond$ w/o All Parts &58.54&55.16&51.14&54.95&$\downarrow$7.29 \\
            \midrule
            \midrule
            Mistral (7B) &\textbf{79.85}&\textbf{60.74}&\textbf{66.54}&\textbf{69.04}&\multicolumn{1}{c}{-} \\
            \midrule
            $\diamond$ w P-Random &77.90&60.63&64.27&67.60&$\downarrow$1.44 \\
            $\diamond$ w P-Input &78.79&60.47&62.80&67.35&$\downarrow$1.69 \\
            $\diamond$ w/o Q Filter &78.28&60.64&58.85&65.92&$\downarrow$3.12 \\
            $\diamond$ w/o P\&Q Filter &76.60&60.42&60.14&65.72&$\downarrow$3.32 \\
            \midrule
            $\diamond$ w/o E Trace &78.69&60.18&64.38&67.75&$\downarrow$1.29 \\
            $\diamond$ w/o All Parts  &74.62&59.11&59.27&64.33&$\downarrow$4.71 \\
		\bottomrule
	\end{tabular}
	\caption{Ablation study for MPT (7B), Falcon (7B), and Mistral (7B) trained with \papertitle. Macro F1-scores are exhibited, and \textbf{$\Delta$\textsubscript{All}} indicates score changes.}
        \label{tab:full_ablation_first}
\end{table}

\begin{table}[t]
\setlength{\tabcolsep}{4.4pt}
    \small
	\centering
	\begin{tabular}{l|cccc|l}
	\toprule
        \multirow{1}{*}{\textbf{Models}}&\multicolumn{1}{c}{\textbf{Noun}} &\multicolumn{1}{c}{\textbf{Verb}}&\multicolumn{1}{c}{\textbf{Event}}&\multicolumn{1}{c}{\textbf{All}}&\multicolumn{1}{c}{\textbf{$\Delta$\textsubscript{All}}}\\
            \midrule
            Llama2 (7B) &\textbf{75.81}&\textbf{59.07}&\textbf{68.00}&\textbf{67.63}&\multicolumn{1}{c}{-} \\
            \midrule
            $\diamond$ w P-Random &69.56&58.48&66.04&64.69&$\downarrow$2.94 \\
            $\diamond$ w P-Input &69.92&58.43&66.34&64.90&$\downarrow$2.73\\
            $\diamond$ w/o Q Filter &65.06&56.90&62.70&61.55&$\downarrow$6.08 \\
            $\diamond$ w/o P\&Q Filter &65.79&57.27&54.52&59.19&$\downarrow$8.44 \\
            \midrule
            $\diamond$ w/o E Trace &69.98&58.25&66.27&64.84&$\downarrow$2.79 \\
            $\diamond$ w/o All Parts &66.34&55.72&55.11&59.05&$\downarrow$8.58\\
            \midrule
            \midrule
            Llama2 (13B) &\textbf{80.35}&\textbf{60.58}&\textbf{67.24}&\textbf{69.39} &\multicolumn{1}{c}{-} \\
            \midrule
            $\diamond$ w P-Random &69.73&60.19&59.40&63.11&$\downarrow$6.28 \\
            $\diamond$ w P-Input &78.46&59.18&65.61&67.75&$\downarrow$1.64 \\
            $\diamond$ w/o Q Filter &72.64&60.17&52.10&61.64&$\downarrow$7.75 \\
            $\diamond$ w/o P\&Q Filter &74.83&59.88&52.54&62.42&$\downarrow$6.97 \\
            \midrule
            $\diamond$ w/o E Trace &79.88&60.46&65.46&68.60&$\downarrow$0.79 \\
            $\diamond$ w/o All Parts &76.05&60.36&59.59&65.33&$\downarrow$4.06 \\
		\bottomrule
	\end{tabular}
	\caption{Ablation study for Llama2 (7B) and Llama2 (13B) trained with \papertitle. Macro F1-scores are exhibited, and \textbf{$\Delta$\textsubscript{All}} indicates score changes.}
        \label{tab:full_ablation_second}
\end{table}

\subsection{Full Results of Ablation Study}
\label{app:full_ablation_study}
Here, we present the full ablation study results of all LLMs trained with our framework \papertitle in \cref{tab:full_ablation_first,tab:full_ablation_second}.

\begin{table}[t]
\setlength{\tabcolsep}{5.2pt}
    \small
	\centering
	\begin{tabular}{l|cccc}
	\toprule
        \multirow{1}{*}{\textbf{Models}}&\multicolumn{1}{c}{\textbf{Acc}} &\multicolumn{1}{c}{\textbf{Ma-F1}}&\multicolumn{1}{c}{\textbf{$\Delta$\textsubscript{Acc}}}&\multicolumn{1}{c}{\textbf{$\Delta$\textsubscript{Ma-F1}}}\\ 
            \midrule
            \multicolumn{5}{l}{\textbf{Fine-tuned on AbsPyramid}} \\
            \midrule
            MPT (7B) &60.42&60.27&-&- \\
            Falcon (7B) &64.22&64.22&-&- \\
            Mistral (7B) &64.81&64.78&-&-  \\
		  Llama2 (7B) &62.40&62.13&-&- \\
            Llama2 (13B) &64.28&64.25&-&- \\
            \midrule
            \multicolumn{1}{l}{\textbf{Direct Injection}}\\
            \midrule
            MPT (7B) &63.97&54.35&$\uparrow$3.55&$\downarrow$5.92 \\
            Falcon (7B) &61.46&55.60&$\downarrow$2.76&$\downarrow$8.62 \\
            Mistral (7B) &70.81&65.26&$\uparrow$6.00&$\uparrow$0.48 \\
		  Llama2 (7B) &69.87&65.92&$\uparrow$7.47&$\uparrow$3.79 \\
            Llama2 (13B) &72.35&68.52&$\uparrow$8.07&$\uparrow$4.27 \\
            \midrule
            \multicolumn{1}{l}{\textbf{AbsInstruct}}\\
            \midrule
            MPT (7B) &71.32&67.55&$\uparrow$10.90&$\uparrow$7.28 \\
            Falcon (7B) &67.82&65.94&$\uparrow$3.60&$\uparrow$1.72 \\
            Mistral (7B) &\textbf{78.21}&\textbf{76.65}&$\uparrow$13.40&$\uparrow$11.87 \\
		  Llama2 (7B) &76.58&75.51&$\uparrow$\textbf{14.18}&$\uparrow$\textbf{13.38} \\
            Llama2 (13B) &77.07&75.44&$\uparrow$12.79&$\uparrow$11.19 \\
		\bottomrule
	\end{tabular}
	\caption{The out-of-domain performance on the AbstractATOMIC dataset. \textbf{$\Delta$\textsubscript{Acc}} and \textbf{$\Delta$\textsubscript{Ma-F1}} mean improvements compared to LLMs fine-tuned on \abspyramid.}
    \label{tab:full_abstract_atomic}
\end{table}

\begin{table}[t]
\setlength{\tabcolsep}{5.2pt}
    \small
	\centering
	\begin{tabular}{l|ccll}
	\toprule
        \multirow{1}{*}{\textbf{Models}}&\multicolumn{1}{c}{\textbf{Acc}} &\multicolumn{1}{c}{\textbf{Ma-F1}}&\multicolumn{1}{c}{\textbf{$\Delta$\textsubscript{Acc}}}&\multicolumn{1}{c}{\textbf{$\Delta$\textsubscript{Ma-F1}}}\\ 
            \midrule
            \multicolumn{5}{l}{\textbf{Fine-tuned on AbsPyramid}} \\
            \midrule
            MPT (7B) &80.38&71.47&\multicolumn{1}{c}{-}&\multicolumn{1}{c}{-} \\
            Falcon (7B) &67.55&63.82&\multicolumn{1}{c}{-}&\multicolumn{1}{c}{-} \\
            Mistral (7B) &79.32&72.66&\multicolumn{1}{c}{-}&\multicolumn{1}{c}{-}  \\
		  Llama2 (7B) &78.69&71.07&\multicolumn{1}{c}{-}&\multicolumn{1}{c}{-} \\
            Llama2 (13B) &82.11&71.25&\multicolumn{1}{c}{-}&\multicolumn{1}{c}{-} \\
            \midrule
            \multicolumn{1}{l}{\textbf{Direct Injection}}\\
            \midrule
            MPT (7B) &78.79&56.69&$\downarrow$1.59&$\downarrow$14.78 \\
            Falcon (7B) &47.54&47.12&$\downarrow$20.01&$\downarrow$16.70 \\
            Mistral (7B) &85.34&74.55&$\uparrow$6.02&$\uparrow$1.89 \\
		  Llama2 (7B)&84.29&74.00&$\uparrow$5.60&$\uparrow$2.93 \\
            Llama2 (13B) &85.51&76.27&$\uparrow$3.40&$\uparrow$5.02 \\
            \midrule
            \multicolumn{1}{l}{\textbf{AbsInstruct}}\\
            \midrule
            MPT (7B)&79.57&70.70&$\downarrow$0.81&$\downarrow$0.77\\
            Falcon (7B)&76.19&69.97&$\uparrow$\textbf{8.64}&$\uparrow$6.15\\
            Mistral (7B)&86.61&77.80&$\uparrow$7.29&$\uparrow$5.14\\
            Llama2 (7B)&84.31&78.76&$\uparrow$5.62&$\uparrow$7.69\\
            Llama2 (13B)&\textbf{87.11}&\textbf{79.89}&$\uparrow$5.00&$\uparrow$\textbf{8.64}\\
		\bottomrule
	\end{tabular}
	\caption{The out-of-domain performance on the Levy/Holt dataset. \textbf{$\Delta$\textsubscript{Acc}} and \textbf{$\Delta$\textsubscript{Ma-F1}} mean improvements compared to LLMs fine-tuned on \abspyramid.}
    \label{tab:full_levy_holt}
\end{table}

\begin{table}[t]
    \small
    \setlength{\tabcolsep}{5pt}
	\centering
	\begin{tabular}{l|cccc}
	    \toprule
		\multirow{2.6}{*}{\textbf{Models}} & \multicolumn{2}{c}{Head Event} & \multicolumn{2}{c}{Exp. Trace} \\ \cmidrule(lr){2-3} \cmidrule(lr){4-5}
  &\textbf{Avg.}&\textbf{Uni.} &\textbf{Avg.}&\textbf{Uni.}\\
            \midrule
            MPT (7B) &0.164&96.00&0.272&96.17 \\
		Falcon (7B) &0.164&96.17&0.276&96.33 \\
            Mistral (7B) &0.157&96.33&0.258&96.83 \\
		  Llama2 (7B) &0.161&96.17&0.261&96.83 \\
            Llama2 (13B) &0.161&96.00&0.256&97.00 \\
		\bottomrule
	\end{tabular}
	\caption{Analysis of diversity of examples collected by our framework when the diversity filter is removed. We list the average ROUGE-L similarity between every pair of samples and the percentage of unique examples.}
        \label{tab:avg_rouge}
\end{table}

\begin{table}[t]
\setlength{\tabcolsep}{5pt}
    \small
	\centering
	\begin{tabular}{l|cccc|c}
	\toprule
        \multirow{1}{*}{\textbf{Models}}&\multicolumn{1}{c}{\textbf{Noun}} &\multicolumn{1}{c}{\textbf{Verb}}&\multicolumn{1}{c}{\textbf{Event}}&\multicolumn{1}{c}{\textbf{All}}&\multicolumn{1}{|c}{\textbf{$\Delta$\textsubscript{All}}}\\
            \midrule
MPT (7B)&70.27&58.40&64.04&64.24&$\downarrow$0.65\\
Falcon (7B)&66.78&55.88&64.10&62.25&$\uparrow$0.01\\
Mistral (7B)&80.05&60.78&67.08&69.31&$\uparrow$0.27\\
Llama2 (7B)&74.35&59.44&67.27&67.02&$\downarrow$0.61\\
Llama2 (13B)&80.49&60.51&66.92&69.31&$\downarrow$0.08\\
		\bottomrule
	\end{tabular}
	\caption{The performance of ablating the diversity filter. We only see fluctuations due to the high diversity of examples, even without the diversity filter.}
    \label{tab:performance_without_diversity}
\end{table}

\begin{table}[t]
\setlength{\tabcolsep}{5pt}
    \small
	\centering
	\begin{tabular}{l|cccc|c}
	\toprule
        \multirow{1}{*}{\textbf{Models}}&\multicolumn{1}{c}{\textbf{Noun}} &\multicolumn{1}{c}{\textbf{Verb}}&\multicolumn{1}{c}{\textbf{Event}}&\multicolumn{1}{c|}{\textbf{All}}&\multicolumn{1}{c}{\textbf{$\Delta$\textsubscript{All}}}\\ 
            \midrule
            MPT (7B)&70.73&58.64&65.74&65.04&$\uparrow$0.15\\
            Falcon (7B)&67.89&57.49&61.80&62.39&$\uparrow$0.15\\
            Mistral (7B)&76.52&60.93&67.90&68.45&$\downarrow$0.59\\
		  Llama2 (7B)&74.22&59.36&66.84&66.81&$\downarrow$0.82\\
            Llama2 (13B)&78.48&59.79&68.14&68.80&$\downarrow$0.59\\
		\bottomrule
	\end{tabular}
	\caption{Macro F1-scores with ChatGPT as the source of explanation traces. \textbf{$\Delta$\textsubscript{All}} means score changes compared to that with GPT4 as the source.}
    \label{tab:chatgpt_rationale_score}
\end{table}

\subsection{Study of Diversity Filter}
\label{app:diversity_filter_study}
In this appendix, we study the role of diversity filters in our framework \papertitle. Here, we remove the diversity filter and analyze the performance of the ablated framework. 

First, we inspect the diversity of examples collected by the ablated framework. We compute the average ROUGE-L similarity between the head events and between explanation traces. From the \cref{tab:avg_rouge}, we can see that the average ROUGE-L similarities are no more than 0.2 for head events and 0.3 for explanation traces. Meanwhile, we also compute the proportion of unique head events and explanation traces based on ROUGE-L, following previous work~\cite{wang2023self}. A head event $x$ is unique if $Rouge_L(C, x) \leq 0.7$, where $C$ is other head events collected by our framework. We apply the same criterion to identify unique data for explanation traces. From \cref{tab:avg_rouge}, we can see that more than 96\% of head events and explanation traces are unique. These findings of average ROUGE-L and uniqueness percentages demonstrate that our dataset can collect quite diverse examples even without the diversity filter.

Then, we test the performance of the \papertitle framework without the diversity filter, shown in \cref{tab:performance_without_diversity}. We can observe that the performance of all LLMs varies slightly. While we add a filter in our framework to guarantee the diversity of collected examples, our study verifies that the data collected by the ablated framework is already highly diverse.

\subsection{Full Results of Out-of-Domain Evaluation}
\label{app:full_ood_results}
As we only plot the Macro F1-scores in \cref{fig:abstract_atomic}, we provide the full results on the AbstractATOMIC dataset across all metrics in \cref{tab:full_abstract_atomic}. Meanwhile, we provide the results of all LLMs on the Levy/Holt dataset in \cref{tab:full_levy_holt}.

\subsection{Full Results of ChatGPT Rationales}
\label{app:full_chatgpt_rationale}
As we only plot the Macro F1-score on the whole test set of \abspyramid in \cref{fig:chatgpt_rationale_score}, we provide the full results on each entailment relation of \abspyramid in \cref{tab:chatgpt_rationale_score}.

\begin{figure}[t]
    \centering
    \includegraphics[width=0.98\columnwidth]{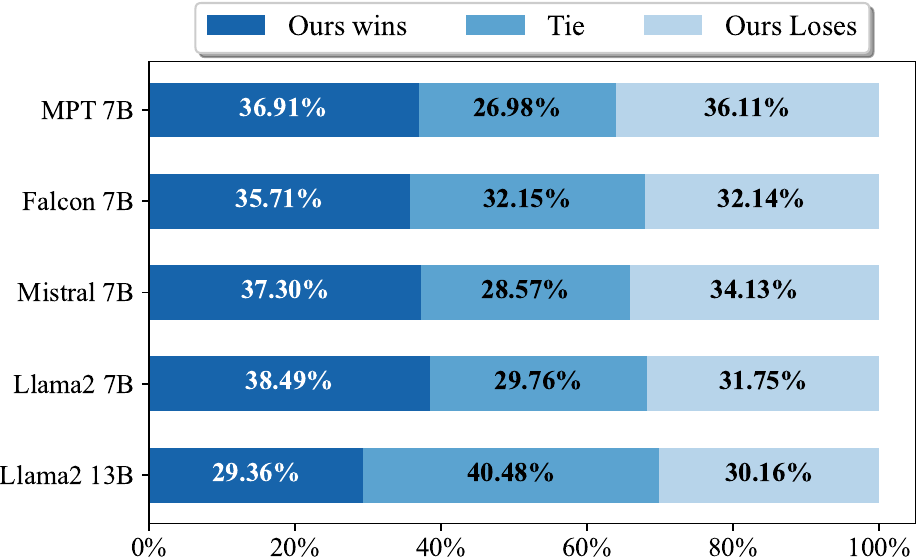}
    \caption{Human preference evaluation, comparing our framework to LLMs trained on Alpaca across 252 test
prompts from \selfinstruct.}
    \label{fig:human_eval}
\end{figure}

\subsection{Human Instruction Following}
\label{app:instruct_human_evaluation}
As previous works~\cite{ouyang2022training, zhao2023survey} suggest a disparity between NLP tasks and human requests, we manually evaluate our framework on the 252 expert-curated instructions of \selfinstruct~\cite{wang2023self} to better understand the alignment with human values. 
Similar to our evaluation on SuperNI, we consider LLMs trained on Alpaca as baselines. Three expert annotators are asked to compare responses from our framework to the baseline and label which one they prefer. We provide annotation details in \cref{app:human_instruction_following}. Our human preference annotation results are plotted in \cref{fig:human_eval}. We observe that a significant portion of prompts are labeled as ``Tie.'' Also, the winning rates appear to be comparable to, or even exceed, those of baselines preferred. These findings again manifest that our framework preserves LLMs' general capabilities while enhancing their abstraction ability.

\subsection{Case Study and Error Analysis}
\label{app:case_study}
In this section, we provide two examples with responses from our framework and the ``Direct Injection'' baseline. Here, the LLM we use is Llama2 (7B). As shown in \cref{tab:case_study_our_method}, we can see that the baseline model cannot generate the correct answers. In contrast, our framework can explain and compare meanings of the instances and concepts in these examples and then give correct labels.

Then, we also provide an example in \cref{tab:error_analysis_our_method}, where our framework (Llama2 7B) gives wrong predictions. We can see that the model makes wrong conclusions while it explains the instance and concept correctly.

\section{Study of Filtered Examples}
\label{app:filtered_example}
In this appendix, we provide a few examples discarded by each quality filter to show their effectiveness. First, we provide two examples that are discarded by the prediction filter in \cref{tab:case_study_prediction}. In these cases, GPT4 generates wrong explanations and predicts wrong labels for these examples. For example, GPT4 considers ``marriage'' as ``a social and legal institution,'' while it refers to ``the legal relationship between a husband and wife'' in the current context. Meanwhile, GPT4 considers ``concession'' as a ``grant or a contract,'' while it refers to ``a right or an advantage that is given to a group of people, an organization, etc.'' We can find that GPT4 can generate wrong explanations and in turn, get wrong answers. 

Then, we also list two examples discarded by the keyword filter in \cref{tab:case_study_keyword}. While GPT4 predicts the correct labels for these examples, it fails to explain the correct instance or concepts. For example, GPT4 explains ``dad-mouthing'' in the first example instead of the correct instance ``quoted.'' In the second example, GPT4 changes the original sentence to ``the things that happen to PersonY,'' which is indeed a fragment and lacks a complete sentence structure.

\begin{table}[t]
\centering
\small
\begin{tabular}{p{0.96\columnwidth}}
\toprule
\multicolumn{1}{c}{\textbf{Example \#1}}
\\ \midrule
\textbf{Entailment Relation:} \nounrelation, \textbf{Label:} 1
 \\
 \textbf{Instance:} walls, \textbf{Concept:} architectural element
 \\ \midrule
\textbf{Head Event:} The \textbf{walls} are white
\\
\textbf{Tail Event:} The \textbf{architectural elements} are white 
 \\ \midrule
 \textbf{Our Framework Response:}
 \\ 
 Step 1: The original word ``walls'' refers to the structural elements that divide or enclose an area, typically made of a solid material such as brick, stone, or wood. Meanwhile, the new word ``architectural element'' refers to any part of a building that is designed and constructed by an architect or engineer, such as walls, windows, doors, stairs, and roofs.
 \\
Step 2: Yes, the meaning of ``architectural element'' encompasses ``walls.''
\\
\textbf{Predicted Label:} 1
\\ \midrule
\textbf{Baseline Response:} No. \textbf{Predicted Label:} 0 \\
 \bottomrule \toprule
 \multicolumn{1}{c}{\textbf{Example \#2}}
\\ \midrule
\textbf{Entailment Relation:} \nounrelation, \textbf{Label:} 1
 \\
 \textbf{Instance:} routine investigation controllers, \textbf{Concept:} controllers
 \\ \midrule
\textbf{Head Event:} A \textbf{routine investigation controllers} are usually given any needed training.
\\
\textbf{Tail Event:} A \textbf{controllers} are usually given any needed training. 
 \\ \midrule
 \textbf{Our Framework Response:}
 \\
 Step 1: The original word ``routine investigation controllers'' refers to individuals or entities that are responsible for conducting routine investigations. Meanwhile, the new word ``controllers'' refers to individuals or entities that have control over something.
 \\
Step 2: Yes, the meaning of ``controllers'' encompasses ``routine investigation controllers.''
\\
\textbf{Predicted Label:} 1
\\ \midrule
 \textbf{Baseline Response:} No. \textbf{Predicted Label}: 0
\\
\bottomrule
\end{tabular}
\caption{Two examples demonstrating that our framework can deliver right answers with explanations, whereas the baseline ``Direct Injection'' falls short in providing correct answers. The LLM we use here is Llama2 (7B).}
\label{tab:case_study_our_method}
\end{table}

\begin{table}[t]
\centering
\small
\begin{tabular}{p{0.96\columnwidth}}
\toprule
\multicolumn{1}{c}{\textbf{Example \#1}}
\\ \midrule
\textbf{Entailment Relation:} \verbrelation, \textbf{Label:} 0
 \\
 \textbf{Instance:} remove, \textbf{Concept:} withdraw
 \\ \midrule
\textbf{Head Event:} The asphalt shingles should be \textbf{removed} first.
\\
\textbf{Tail Event:} The asphalt shingles should be \textbf{withdrawed} first. 
 \\ \midrule
 \textbf{Our Framework Response:}
 \\ 
 Step 1: The original word ``removed'' means to take something away or out of a place. Meanwhile, The new word ``withdraw'' means to take something back or away from a place.
 \\
Step 2: Yes, the meaning of ``withdraw'' encompasses ``removed.''
\\
\textbf{Predicted Label:} 1
\\
\bottomrule
\end{tabular}
\caption{An example showing some errors made by our framework. The LLM we use here is Llama2 (7B).}
\label{tab:error_analysis_our_method}
\end{table}

\begin{table*}[h]
    \small
	\centering
	\begin{tabular}{l|l||cc|cc|cc|cc}
	\toprule
        \multirow{2}{*}{\textbf{Methods}}&\multirow{2}{*}{\textbf{Backbone}}&\multicolumn{2}{c|}{\textbf{Noun}} &\multicolumn{2}{c|}{\textbf{Verb}}&\multicolumn{2}{c|}{\textbf{Event}}&\multicolumn{2}{c}{\textbf{All}}\\ 
	&&\textbf{Acc} &\textbf{Ma-F1} &\textbf{Acc}&\textbf{Ma-F1} & \textbf{Acc}&\textbf{Ma-F1}&\textbf{Acc}&\textbf{Ma-F1} \\
            \midrule
            \textbf{Random} & \multicolumn{1}{|c||}{-} & 50.00 & 49.67 & 50.00 & 49.97 & 50.00 & 49.01 & 50.00 & 49.55 \\
            \midrule
            \multirow{5}{*}{\textbf{Alpaca (10-shot)}}
            &MPT (7B) &44.70&35.68&49.23&37.94&65.53&44.75&53.16&39.45 \\
            &Falcon (7B) &59.98&54.61&55.60&55.45&63.75&45.02&59.77&51.69 \\
            &Mistral (7B) &74.81&73.19&58.76&58.20&65.69&58.78&66.42&63.39 \\
		  &Llama2 (7B) &63.62&63.55&54.86&52.23&69.44&60.60&62.64&58.79 \\
            &Llama2 (13B) &75.13&72.41&58.68&58.66&66.97&61.99&66.93&64.35 \\
            \midrule
            \multirow{5}{*}{\textbf{Direct Injection}}
            &MPT (7B) &64.72&64.14&54.25&52.65&52.14&52.03&57.04&56.27 \\
            &Falcon (7B) &63.16&58.47&54.47&54.31&51.72&51.67&56.45&54.82 \\
            &Mistral (7B) &74.30&74.18&59.56&59.24&59.63&59.05&64.49&64.16 \\
		  &Llama2 (7B) &67.30&66.57&55.24&54.26&57.51&57.49&60.02&59.44 \\
            &Llama2 (13B) &74.55&73.87&59.89&59.74&61.36&60.58&65.27&64.73 \\
            \midrule
            \multirow{5}{*}{\textbf{AbsInstruct}}
            &MPT (7B) &72.27&71.94&59.13&59.11&69.18&66.76&66.86&65.94 \\
            &Falcon (7B) &67.53&67.14&56.42&55.56&68.54&63.49&64.17&62.06 \\
            &Mistral (7B) &\underline{79.98}&\underline{79.45}&59.98&\underline{59.92}&71.96&67.54&\underline{70.64}&\underline{68.97} \\
		  &Llama2 (7B) &77.68&76.69&\underline{60.06}&59.72&\textbf{72.86}&\textbf{68.28}&70.20&68.23 \\
            &Llama2 (13B) &\textbf{80.33}&\textbf{79.73}&\textbf{60.25}&\textbf{60.24}&\underline{72.68}&\underline{68.00}&\textbf{71.09}&\textbf{69.32} \\
		\bottomrule
	\end{tabular}
	\caption{Performance of \papertitle and baselines on the validation set of \abspyramid. \textbf{Acc} and \textbf{Ma-F1} represent Accuracy and Macro F1-score, respectively. We bold the best score and underline the second-best score.}
 \label{tab:main_eval_validation}
\end{table*}

\begin{table}[t!]
\centering
\small
\begin{tabular}{p{0.96\columnwidth}}
\toprule
\multicolumn{1}{c}{\textbf{Example \#1}}
\\ \midrule
 \textbf{Entailment Relation:} \nounrelation
 \\
 \textbf{Instance:} marriage, \textbf{Concept:} marriage bond
 \\ \midrule
\textbf{Head Event:} Their \textbf{marriage} declared invalid. 
\\
\textbf{Tail Event:} Their \textbf{marriage bond} declared invalid. 
 \\ \midrule
 \textbf{GPT4 Rationale:}
 \newline
Step 1: The original word ``marriage'' broadly refers to a social and legal institution, which encompasses a wide range of practices, ceremonies, rights, and significances depending on the culture and context.
\newline
Step 2: The new word ``marriage bond'' typically refers to the specific connection, commitment, or tie that exists between the married individuals.
\newline
Step 3: No.
\\ \midrule
 \textbf{Ground Truth:} Positive, \textbf{Prediction:} Negative.
 \\ \bottomrule \toprule
\multicolumn{1}{c}{\textbf{Example \#2}}
\\ \midrule
 \textbf{Entailment Relation:} \eventrelation
 \\
  \textbf{Instance:} It gives PeopleX the right to fish, \textbf{Concept:} fishing concession
  \\ \midrule
\textbf{Head Event:} It gives PeopleX the right to fish
 \\
 \textbf{Tail Event:} fishing concession
 \\ \midrule
 \textbf{GPT4 Rationale:} 
 \newline
 Step 1: The meaning of the sentence ``It gives PeopleX the right to fish'' is that a certain group, PeopleX, is granted the permission or legal authority to engage in fishing activities.
 \newline
 Step 2: The meaning of the abstract description ``fishing concession'' generally refers to a grant or contract giving an individual or group the right to operate a fishing activity in a specified area or under certain conditions.
 \newline
 Step 3: Yes.
 \\ \midrule
 \textbf{Ground Truth:} Negative, \textbf{Prediction:} Positive.
 \\
\bottomrule
\end{tabular}
\caption{Two examples filtered by the Prediction Filter. GPT4 might generate wrong explanations with wrong predicted labels.}
\label{tab:case_study_prediction}
\end{table}

\begin{table}[t]
\centering
\small
\begin{tabular}{p{0.96\columnwidth}}
\toprule
\multicolumn{1}{c}{\textbf{Example \#1}}
\\ \midrule
\textbf{Entailment Relation:} \verbrelation, \textbf{Label:} Negative
 \\
 \textbf{Instance:} quoted, \textbf{Concept:} intercommunicated
 \\ \midrule
\textbf{Head Event:} PersonX is \textbf{quoted} bad-mouthing teammates
\\
\textbf{Tail Event:} PersonX is \textbf{intercommunicated} bad-mouthing teammates
 \\ \midrule
 \textbf{GPT4 Rationale:} 
 \newline
 Step 1: ``Bad-mouthing'' means to speak negatively or disparagingly about someone.
 \newline 
 Step 2: ``Intercommunicate'' means to communicate with each other, implying a mutual exchange of information.
 \newline
 Step 3: No.
 \\
\bottomrule \toprule
 \multicolumn{1}{c}{\textbf{Example \#2}}
\\ \midrule
 \textbf{Entailment Relation:} \eventrelation, \textbf{Label:} Positive
 \\
 \textbf{Instance:} The things happen to PersonY, \textbf{Concept:} life circumstances
 \\ \midrule
\textbf{Head Event:} The things happen to PersonY
\\
\textbf{Tail Event:} life circumstances
 \\ \midrule
 \textbf{GPT4 Rationale:} 
 \newline
 Step 1: The sentence ``the things that happen to PersonY'' refers to any events, situations, or experiences that occur in the life of an individual named PersonY.
 \newline 
 Step 2: The term ``life circumstances'' broadly refers to the conditions or events that collectively define an individual's situation in life at any given time, encompassing a range of personal, social, economic, and environmental factors that affect the individual.
 \newline
 Step 3: Yes.
\\
\bottomrule
\end{tabular}
\caption{Two examples filtered by the Keyword Filter. GPT4 might generate explanations of wrong words and rewrite the instance or concept.}
\label{tab:case_study_keyword}
\end{table}

\end{document}